\documentclass[journal]{IEEEtran}

\ifCLASSINFOpdf
  \usepackage[pdftex]{graphicx}
  \usepackage{epsfig}
\else
  \usepackage[dvips]{graphicx}
\fi

\usepackage{amsmath}
\usepackage{bm}

\usepackage{array}
\usepackage[numbers,sort&compress]{natbib}
\usepackage{color}
\usepackage{amsfonts}       
\usepackage{algorithm}
\usepackage{algorithmic}
\usepackage{multirow}

\ifCLASSOPTIONcompsoc
  \usepackage[caption=false,font=normalsize,labelfont=sf,textfont=sf]{subfig}
\else
  \usepackage[caption=false,font=footnotesize]{subfig}
\fi

\begin{document}

\title{Dynamic Ensemble Bayesian Filter for Robust Control of a Human Brain-machine Interface }

\author{
Yu~Qi,
Xinyun Zhu,
Kedi Xu,
Feixiao Ren,
Hongjie Jiang,
Junming Zhu, \\
Jianmin Zhang, 
Gang Pan,
Yueming Wang$^{*}$
\thanks{Yu Qi is with the MOE Frontier Science Center for Brain Science and Brain-machine Integration, and the Department of Neurobiology, Affiliated Mental Health Center \& Hangzhou Seventh People's Hospital, Zhejiang University School of Medicine, Hangzhou, China. Xinyun Zhu, Feixiao Ren, and Yueming Wang are with the Qiushi Academy for Advanced Studies, Zhejiang University, Hangzhou, China. Kedi Xu is with the Zhejiang Lab, and the Qiushi Academy for Advanced Studies, Zhejiang University, Hangzhou, China. Hongjie Jiang, Junming Zhu, and Jianmin Zhang are with the Second Affiliated Hospital of Zhejiang University School of Medicine, Hangzhou, China. Gang Pan is with the College of Computer Science and Technology, Zhejiang University, Hangzhou, China.
The corresponding author is Yueming Wang (ymingwang@zju.edu.cn).
}}

\maketitle

\IEEEpeerreviewmaketitle

\begin{abstract}

\textit{Objective:} Brain-machine interfaces (BMIs) aim to provide direct brain control of devices such as prostheses and computer cursors, which have demonstrated great potential for mobility restoration. One major limitation of current BMIs lies in the unstable performance in online control due to the variability of neural signals, which seriously hinders the clinical availability of BMIs. 
\textit{Method:}
{
To deal with the neural variability in online BMI control, we propose a dynamic ensemble Bayesian filter (DyEnsemble). DyEnsemble extends Bayesian filters with a dynamic measurement model, which adjusts its parameters in time adaptively with neural changes. This is achieved by learning a pool of candidate functions and dynamically weighting and assembling them according to neural signals.
In this way, DyEnsemble copes with variability in signals and improves the robustness in online control. 
}
\textit{Results:}
Online BMI experiments with a human participant demonstrate that, compared with velocity Kalman filter, DyEnsemble significantly improves the control accuracy (increases the success rate by 13.9\% and reduces the reach time by 13.5\% in the random target pursuit task) and {robustness (performs more stably over different experiment days).}
\textit{Conclusion:}
Our results demonstrate the superiority of DyEnsemble in online BMI control. 
\textit{Significance:}
DyEnsemble frames a novel and flexible framework for robust neural decoding, which is beneficial to different neural decoding applications. 

\end{abstract}

\section{Introduction}

In recent decades, brain-machine interfaces (BMIs) have been demonstrated with great potential of restoring mobility for humans paralyzed by trauma to the neural system, such as spinal cord injuries \cite{hochberg2006neuronal,hochberg2012reach,bouton2016restoring,pandarinath2017high, pan2018rapid}. Neural activities, recorded intracortically by microelectrodes placed on the primary motor cortex or parietal cortical areas \cite{aflalo2015decoding}, could be translated into control signals to external assistive devices, such as computer cursors and robotic limbs. 

Recent advances in BMIs have demonstrated the potential for high-performance communication and control. Much effort has been made to establish real-time motor BMIs to link brain and prosthetic limbs, including studies on the brain control of arms and hands for reaching and grasping \cite{hochberg2012reach}, the motion of upper and lower limbs \cite{Nicolelis2009walking}, and the degrees of freedom for arm movements \cite{collinger2013high-performance}. Decoding methods, the mathematical models transforming neuronal signals to behavioral parameters to control external devices, play an important role in BMI. Since Georgopoulos proposed the classical population encoding mechanism \cite{Georgopoulos1989populationvector}, linear decoders have been leading the translation of neural activity into movements, including Wiener filter \cite{Nicolelis2009walking}, population vectors (PV) \cite{georgopoulos1986neuronal}, optimal linear estimation (OLE) \cite{wodlinger2015ten-dimensional}, and Kalman filter \cite{Nicolelis2004Human1st}. Nonlinear decoders, which are capable of capturing more complex relationships and more noise robustly, obtained superior decoding performance in offline analysis { \cite{li2021robust, ahmadi2021robust}}. There is also evidence that nonlinear methods, especially those involving artificial neural networks, demonstrated higher performance in certain motion tasks \cite{Shenoy2012RNN, li2021robust}, compared with their linear counterparts. Nevertheless, studies suggested that it is not necessarily true that improvements in offline reconstruction will translate to improvements in online control, considering the participant's learning and compensation ability \cite{koyama2010comparison, dangi2013design}. Therefore, although nonlinear models might construct more powerful decoders, linear decoders such as OLE and Kalman filter are still the most popular choice for online control. 

{
One critical barrier to clinical available BMI applications is the unstable performance due to the variability in neural signals. That is, individual neurons can exhibit significant variability in firing properties across different trials in a motor task \cite{sussillo2016making, suway2018temporally, even2017augmenting}. In closed-loop BMI control, the variability of neural signals can be even more complicated with the influence of real-time feedback. Because a user will update the control strategy when the output of the decoder does not match the intent, i.e., user's adaptation to the decoder, which can apparently change neuronal tuning \cite{taylor2002direct}. Besides, different conditions of neural feedbacks (e.g., the speed of cursor) can induce variability in neural response and encoding \cite{churchland2007temporal, even2017augmenting, churchland2006preparatory}. 
If the variability is not properly addressed, it may lead to mistakes in intention interpretation, thus seriously hindering the availability of BMI systems. 
}
To improve the robustness of neural decoding, some studies transform the original neural activities to a lower-dimensional space using dimension reduction methods such as principal component analysis or factor analysis \cite{santhanam2009factor, degenhart2020stabilization}, where the effect of variability can be suppressed. {Other ways to deal with the variability include recording more neurons to mitigate non-stationarity of single neurons \cite{degenhart2020stabilization, carmena2005stable}, or using larger datasets and more complex decoders (usually nonlinear) to improve the robustness against variability \cite{sussillo2016making}. However, the unstable performance brought by neural variability is still a challenging problem that has not been well addressed. }

{
This study focuses on dealing with the variability of neural encoding associated with brain states in closed-loop BMI systems, which leads to shifts in the functional relationship between neural signals and kinematics \cite{churchland2007temporal, even2017augmenting, churchland2006preparatory}. To this end, this study proposes an dynamic ensemble Bayesian filter (DyEnsemble) for robust BMI control. 
Unlike classical Bayesian filters which use a fixed model, DyEnsemble learns a pool of models that contains diverse abilities in describing the neural encoding functions in different time slots. Then in each time slot, it seeks a proper ensemble model by dynamically weighting and assembling the models in the pool according to the neural signals. In this way, it improves the robustness against variability. 
}
To dynamically assemble a neural decoder along with signals, there are three main issues to address: 
1) how to dynamically weight and assemble models in an online process; 
2) how to construct the model pool to cover the variability in neural signals;
3) how to sequentially estimate the kinematic intentions with the DyEnsemble model. 
For 1), we propose to adaptively tune the weights of models by how much the observation neural signals support each model (i.e. the likelihood of model), and assemble the models according to the Bayesian model averaging algorithm. For 2), we construct multiple models of linear, polynomial, and neural networks to cover the functional variability in neural encoding. For 3), we propose a particle-based solution to simultaneously estimate both the model ensemble and the kinematics recursively in time.

The proposed DyEnsemble model extends Bayesian neural decoders by incorporating Bayesian filtering and Bayesian model averaging in a uniform state-space formulation and a particle-based solution. 
With the dynamic model ensemble process, DyEnsemble can adaptive to changes in neural activities; thus can better cover the neural variability to obtain more robust online BMI control. 
In closed-loop BMI experiments with a human subject, DyEnsemble achieves significant performance improvement compared with the velocity Kalman filter. In the random target pursuit task, DyEnsemble increases the success rate by 13.9\%, decreases the reach time by 13.5\%, { and performs more stably over different experiment days,} which demonstrates superior control accuracy and robustness. Analysis shows that DyEnsemble can adaptively adjust the model weights and incline to proper models along with changes in signals; thus, robust performance can be achieved.

\section{DyEnsemble Bayesian Filter}

\begin{figure*}[h]
\centering
\includegraphics[scale=0.4]{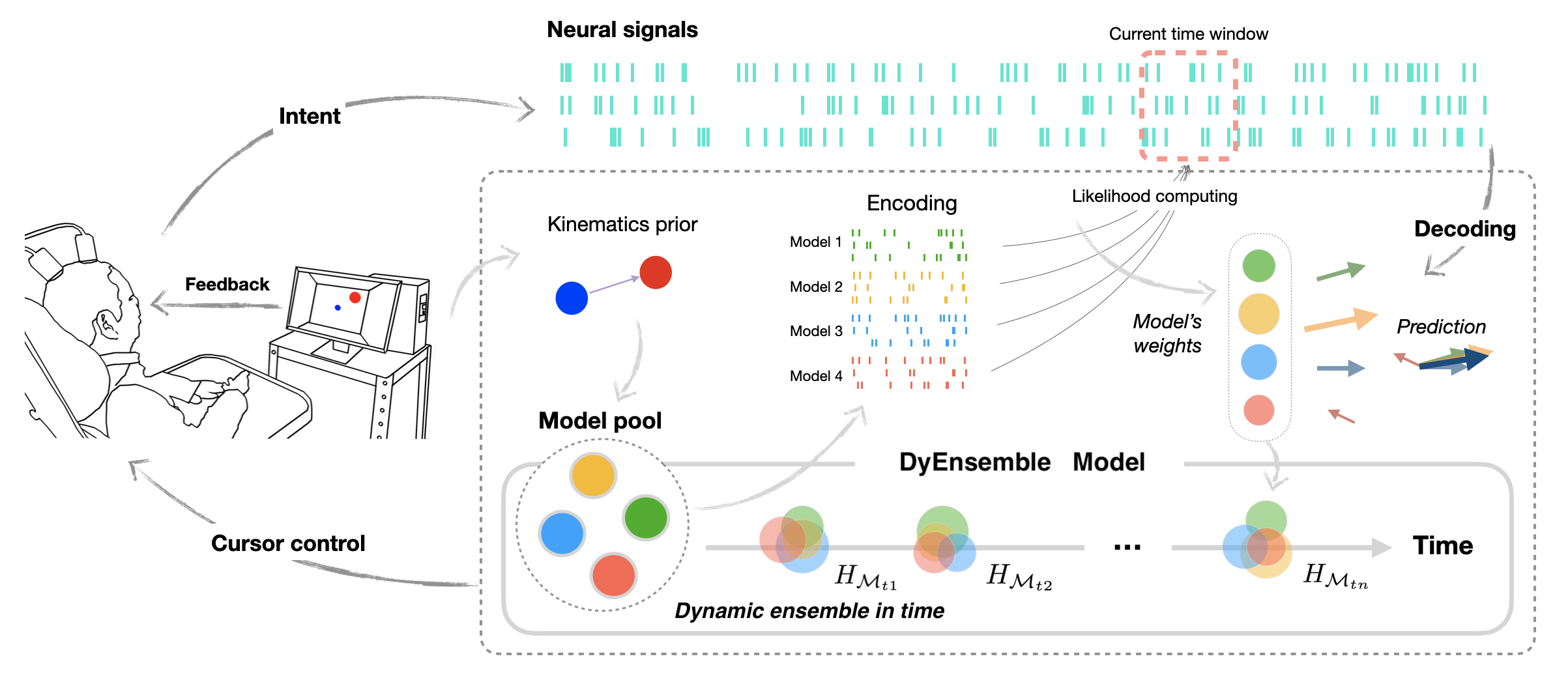}
\caption{The framework of the DyEnsemble decoder. DyEnsemble learns a pool of different models that have diverse abilities in describing the functional mapping between neural signals and velocity in each time slot. To find the proper models dynamically at each time slot, DyEnsemble evaluates the models in the pool by the Bayesian likelihood of their estimated neural signals (encoded from the kinematics prior) to the observed neural signals. Based on this, models with higher likelihood will obtain larger weights in the ensemble. The  assembled decoder is used for predicting the velocity in the current time slot.
}
\label{fig:framework}
\end{figure*}

The framework of DyEnsemble is given in Fig. \ref{fig:framework}. In this section, we firstly present the state-space formulations of DyEnsemble. Then we introduce the model pool construction strategy in this study. After that, we describe the dynamic model ensemble algorithm in the online test stage. Finally, we give the particle-based solution to estimate both the dynamic model and the kinematic states in DyEnsemble.

\subsection{State-space Formulation}

In the neural decoding problem, the objective is to estimate kinematic parameters (state) given neural signals (observation) sequentially in time. The state-space model of DyEnsemble contains two parts: a state transition function and a measurement function. 
The state transition function describes how kinematic parameters evolve over time, and the measurement function describes the functional relationship between kinematics and neural signals.

In the state-space formulation, DyEnsemble extends classical Bayesian filters with a dynamically assembled measurement function. Unlike typical Bayesian filters, which adopt fixed measurement functions, the measurement model in DyEnsemble changes in time based on neural signals, and the model of measurement noise updates accordingly as well. The state-space formulation of DyEnsemble is as follows:
        \begin{flalign}
        \label{equation:state}
        \bm{x}_t=A\bm{x}_{t-1}+\bm{b}+\bm{u}_{t-1}\\
        \label{equation:observation}
         \bm{y}_t = \mathcal{H}_{\mathcal{M}_t}(\bm{x}_t)+\bm{v}_{\mathcal{V}_t}
         \end{flalign}
{where (\ref{equation:state}) and (\ref{equation:observation}) are the state transition and measurement model, respectively. In the equations, $t$ denotes the time step. $\bm{x}_t$ is the kinematic parameters to be estimated, which is a $2 \times 1$ vector of $[v_x,v_y]$, { where $v_x$, $v_y$ represent the velocities in x-axis and y-axis, respectively. }$\bm{y}_t$ is the firing rate of neurons in a 20 ms bin at time $t$, which is a $C \times 1$ vector with $C$ denoting the number of neurons. 

In (\ref{equation:state}), $A$ and $\bm{b}$ are the transition matrix and bias, with $\bm{u}_{t} \sim N(0, \sigma^2_{u})$ is the i.i.d Gaussian transition noise with a zero mean and a variance of $\sigma^2_{u}$. The parameters of $A$ and $\bm{b}$ are estimated with the least square algorithm, $\bm{u}_{t}$ is estimated with the fitting residuals. 
}

In (\ref{equation:observation}), $\mathcal{M} = [m_1, m_2, ... m_q]$ is a set of candidate measurement functions in which $m_k: \mathbb{R}^{n_{x}} \rightarrow \mathbb{R}^{n_{y}}$, where $(k=1, ... , q)$ is a functional mapping that infers the neural signals given kinematic parameters. $n_x$ and $n_y$ are the dimensions of $\bm{x}_t$ and $\bm{y}_t$, respectively. At a time step $t$, the measurement function $\mathcal{H}_{\mathcal{M}_{t}}$ is a weighted combination of functions in $\mathcal{M}$: 

\begin{equation}
\label{equation:model_ensemble}
 \mathcal{H}_{\mathcal{M}_{t}} = \sum_{k=1}^{q}{w_{k, t}m_k(.)}
\end{equation}
{
where the weight $w_{k, t}$ is dynamically updated at each $t$ according to the fitness of functions given neural signals. Accordingly, the parameters of $\mathcal{H}_{\mathcal{M}_{t}}$ changes adaptively with neural signals. 
}

For the measurement noise term $\bm{v}_{\mathcal{V}_t}$, suppose we have a set of $\mathcal{V} = [\widehat{\bm{v}}_1, \widehat{\bm{v}}_2, ... \widehat{\bm{v}}_q]$, where $\widehat{\bm{v}}_k \sim N(0, \sigma^2_{\widehat{\bm{v}}_k})$ is the i.i.d measurement noise corresponding to the $k^{th}$ model in $\mathcal{M}$. The measurement noise model at time $t$ can be defined by:
        \begin{flalign}
        \label{equation:m_noise}
         \bm{v}_{\mathcal{V}_{t}} = \sum_{k=1}^{q}{w_{k, t}\widehat{\bm{v}}_k}.  
         \end{flalign}

\subsection{Model Pool Construction}

Here we introduce how the model set $\mathcal{M}$ is constructed in DyEnsemble. In the measurement model, function $\mathcal{H}_{\mathcal{M}_t}$ transforms kinematics to neural activities, which can be regarded as an ``encoding model" or encoder. 
Therefore, the model pool can be built by enrolling diverse neural encoders. During online BMI control, DyEnsemble can dynamically weight and assemble the neural encoders adaptively according to neural signals.

A classical neural encoder is a linear mapping from kinematics to neural activities according to the tuning curve theory, namely the firing rate of an individual unit is cosine-tuned to the movement directions \cite{georgopoulos1982relations, chase2009bias}. 
Recent studies suggest that the relationship between kinematics and neural activities contains nonlinearity \cite{inoue2018decoding, liang2019deep}. {Study \cite{liang2019deep} compared different neural encoding models, including preferred direction model (PD) \cite{georgopoulos1982relations}, Poisson process velocity tuning model (PPVT) \cite{kao2014information} , generalized linear models (GLM) \cite{gerwinn2010bayesian,truccolo2005point,benjamin2018modern}, and deep learning-based approaches such as multilayer perceptron model (MLP) and recurrent neural network (RNN).} Results demonstrated that deep learning neural encoders show high neural data reproducing performance, and GLM obtained the best performance among the linear models. 

{
To deal with the variability of neural encoding in online BMI control, we propose to enroll diverse types of neural encoders.} In this study, we involve four encoders in the model pool of DyEnsemble: a linear encoder, a polynomial encoder, and two neural network models with different parameter sizes. The details of the encoding models are specified as follows:

\begin{itemize}
\item
\textbf{Linear model.} We adopt the linear Gaussian GLM model $\mathcal{H}_{l}$ to learn a functional mapping from two-dimensional velocity to neural firing rates:
{
        \begin{flalign}
         \mathcal{H}_{l}(\bm{x}_{t}) =  W\bm{x}_{t} + \widehat{\bm{v}}_{l,t}.
         \end{flalign}
         }
The parameter $W$ can be estimated with the least square algorithm, and the noise term $\widehat{\bm{v}}_{l,t} \sim N(0, \sigma^2_{\widehat{\bm{v}}_{l}})$ is estimated with the fitting residuals.
\item
\textbf{Polynomial model.} The polynomial encoder $\mathcal{H}_{p}$ maps the relationship between kinematics and neural activities in a second-order polynomial function:
{
        \begin{flalign}
         \mathcal{H}_{p}(\bm{x}_{t})  =  W_2\bm{x}_{t}^2 + W_1\bm{x}_{t} + \widehat{\bm{v}}_{p,t}. 
         \end{flalign}
         }
The parameters are estimated by ridge regression, and the noise term $\widehat{\bm{v}}_{p,t} \sim N(0, \sigma^2_{\widehat{\bm{v}}_{l}})$ is estimated with the fitting residuals.

\item
\textbf{Neural network models.} Two MLP models with different parameter sizes are employed to learn the encoding functions, namely $\mathcal{H}_{nn_1}$ and $\mathcal{H}_{nn_2}$. For both neural networks, one hidden layer is employed. The neuron number of the input layer is the size of the kinematic parameter vector $\bm{x}_t$ and the neuron number of the output layer is the size of the neural signal $\bm{y}_t$. The neuron number of the hidden layers are set to 30 and 50 respectively for the two neural networks. The parameters are optimized with the Adam algorithm \cite{kingma2014adam} with a learning rate of 0.01 and a weight decay of 1e-4. Early stop strategy is adopted to mitigate the overfitting problem. The noise term $\widehat{\bm{v}}_{nn_1,t} \sim N(0, \sigma^2_{\widehat{\bm{v}}_{nn_1}})$ and $\widehat{\bm{v}}_{nn_2,t} \sim N(0, \sigma^2_{\widehat{\bm{v}}_{nn_2}})$ are estimated with the fitting residuals, respectively.

\end{itemize}

In this work, the model pool is defined by $\mathcal{M} = [\mathcal{H}_{l}, \mathcal{H}_{p}, \mathcal{H}_{nn_1}, \mathcal{H}_{nn_2}]$. Diverse settings can be applied in model pool construction according to different tasks.

\subsection{Dynamic Model Ensemble}

One key problem in DyEnsemble is how to assemble the models dynamically during online BMI control, namely, how to estimate $w_{k,t}$ for each model in $\mathcal{M}$ at each time step $t$. There are four steps for weights calculation and states prediction. The first step is the neural encoding process with the models in the pool. Then the models are weighted according to the likelihood of the encoded neural signals given the observed neural signals at time ${t}$. After that, the multiple models are assembled to obtain the measurement function at time ${t}$ by Bayesian model averaging, and finally $x_{t}$ can be estimated with recursive Bayesian filtering.

\subsubsection{Step 1: Neural encoding with multiple models}

Given the kinematics of time $t-1$, we can predict the prior distribution of kinematics $p(\bm{x}_t|\bm{x}_{t-1})$ with the state transition function. Then using the prior kinematics as the input, we infer the neural signals with the encoding models in the model pool as 
	\begin{equation}
         \bm{\hat{y}}_{k, t} =  m_k(\bm{x}_t)
         \end{equation}
where $m_k \in \mathcal{M} = [\mathcal{H}_{l}, \mathcal{H}_{p}, \mathcal{H}_{nn_1}, \mathcal{H}_{nn_2}]$.

\subsubsection{Step 2: Dynamic model weighting by likelihood of encoding}

With the neural signal inferences by the models in the pool, we can evaluate the effectiveness of each model by how likely the truly observed neural signals at time $t$ appear. Specifically, we compute the likelihood of the observed neural signals at time $t$ with model $m_k$ as $p_k(\bm{y}_t|\bm{y}_{0:t-1})$:
        \begin{equation}
        \label{equation:likelihood}
        p_k(\bm{y}_t|\bm{y}_{0:t-1}) = \int{p_k(\bm{y}_t|\bm{x}_t)p(\bm{x}_t|\bm{y}_{0:t-1})d\bm{x}_t }.
        \end{equation}

The models are weighted according to the likelihood values, which indicate how the observation data support each model. In this way, the weights of models can be dynamically adjusted to cope with neural changes brought by different conditions or non-stationarity in signals. The dynamic weighting process is the key point for improving the adaption ability in DyEnsemble.

\subsubsection{Step 3: Model assembling with Bayesian model averaging}

Having the likelihoods of models at time $t$ at hand, we can assemble the measurement model  $\mathcal{H}_{\mathcal{M}_{t}}$ at $t$ with the Bayesian model averaging algorithm \cite{montgomery2010bayesian, fragoso2018bayesian}, where the weight of each model is assigned by its posterior probability $p( \mathcal{H}_{\mathcal{M}_t} = m_k|\bm{y}_{0:t})$ given incoming neural signals. 

The computation of $p( \mathcal{H}_{\mathcal{M}_t}=m_k|\bm{y}_{0:t})$ contains two parts, including the prior distribution of model $m_k$, and the likelihood associated with model $m_k$: 
        \begin{equation}
        \label{equation:post_m}
        \begin{aligned}
        p( \mathcal{H}_{\mathcal{M}_t} &= m_k|\bm{y}_{0:t})= \\
        & \frac{p( \mathcal{H}_{\mathcal{M}_t} = m_k|\bm{y}_{0:t-1})
        p_k(\bm{y}_t|\bm{y}_{0:t-1})}
        {\sum_{j=1}^q{p( \mathcal{H}_{\mathcal{M}_t} = m_j|\bm{y}_{0:t-1})
        p_j(\bm{y}_t|\bm{y}_{0:t-1})}}
        \end{aligned}
        \end{equation}
where $p( \mathcal{H}_{\mathcal{M}_t} = m_k|\bm{y}_{0:t-1})$ denotes the prior distribution of $m_k$, and $p_k(\bm{y}_t|\bm{y}_{0:t-1})$ is the likelihood of model $m_k$ which we have obtained in Step 2.

For the prior distribution, we define the state transition with a forgetting coefficient as in our previous study \cite{qi2019dynamic}:
        \begin{equation}
        \label{equation:forget}
        p( \mathcal{H}_{\mathcal{M}_t} = m_k|\bm{y}_{0:t-1}) = \frac{p( \mathcal{H}_{\mathcal{M}_{t-1}}  =m_k|\bm{y}_{0:t-1})^\alpha}{\sum_{j=1}^q {p( \mathcal{H}_{\mathcal{M}_{t-1}} = m_j|\bm{y}_{0:t-1})^\alpha}}
        \end{equation}
where the forgetting coefficient $\alpha \in (0,1)$ controls the rate of reducing the impact of historical signals.

\subsubsection{Step 4: Kinematics estimation with recursive Bayesian filtering}

According to the state-space modeling of DyEnsemble in (\ref{equation:state}) and (\ref{equation:observation}), given a sequence of neural signals $\bm{y}_{0:t}$, the posterior of the kinematic parameters $\bm{x}_t$ at time $t$ can be formulated by:
        \begin{equation}
        \label{equation:post_xm}
        {p(\bm{x}_t|\bm{y}_{0:t}) = \sum_{k=1}^q{p(\bm{x}_t| \mathcal{H}_{\mathcal{M}_t} =m_k,\bm{y}_{0:t})
        p(\mathcal{H}_{\mathcal{M}_t} = m_k|\bm{y}_{0:t})}}
        \end{equation}
{where $p(\bm{x}_t|\mathcal{H}_{\mathcal{M}_t} = m_k,\bm{y}_{0:t})$ is the posterior of the state with model $m_k$, which can be estimated sequentially with recursive Bayesian filtering, given $m_k$; and $p(\mathcal{H}_{\mathcal{M}_t} = m_k|\bm{y}_{0:t})$ is the posterior probability of $m_k$, which is obtained in Step 3.} 

From the aspect of modeling, DyEnsemble incorporates Bayesian model averaging with recursive Bayesian filtering process, such that the weights of models can be adaptively adjusted with changes in neural signals. From the perspective of Bayesian filtering, DyEnsemble enforces classical Bayesian filters by accounting for the uncertainty of the measurement model. {By simultaneously estimating both the state $\bm{x}_t$, and the measurement function $\mathcal{H}_{\mathcal{M}_t}$ along with time, DyEnsemble improves the decoder's adaption ability against non-stationary neural signals.}

\subsection{Particle-based State Estimation}

In DyEnsemble, the posterior distribution of state $\bm{x}_t$ can be sequentially estimated using a particle-based algorithm proposed in our previous study of \cite{qi2019dynamic}. Here we briefly introduce the particle-based solution for (\ref{equation:post_xm}).

The particle-based estimation of $p(\bm{x}_t| \mathcal{H}_{\mathcal{M}_t}  = m_k,\bm{y}_{0:t})$ is similar to the classical particle filter with the observation model of $m_k(.)$. Given a set of particles $\{\bm{x}^i_{t}\}_{i=1}^{N_s}$, the posterior distribution can be estimated by the particles and their associated weights: 
\begin{equation}
\label{equation:posterior1}
 p_k(\bm{x}_t| \mathcal{H}_{\mathcal{M}_t}  = m_k,\bm{y}_{0:t}) \approx \sum^{N_s}_{i=1}\omega^i_{k,t}\delta (\bm{x}_t-\bm{x}^i_t)
\end{equation}
where $\delta(.)$ denotes the Dirac delta function, $N_s$ is the particle number, and $\bm{x}_t^i$ is the $i^{th}$ particle at time $t$ with $\omega^i_{k,t}$ being its normalized weight associated with the observation function of $m_k(.)$. 
Sequential importance sampling (SIS) \cite{arulampalam2002tutorial} is employed for recursive estimation, and the weights of particles can be updated by $\omega^i_{k,t}\propto \omega^i_{k,t-1}p_{k}(\bm{y}_t|\bm{x}^i_t)$, with $\sum_{i=1}^{N_s}\omega^i_{k,t}=1$.

Given the model set $\mathcal{M} = [m_1, m_2, ... m_q]$, we can estimate $p( \mathcal{H}_{\mathcal{M}_t} = m_k|\bm{y}_{0:t}) $ in (\ref{equation:post_m}). Specifically, the prior of $\mathcal{H}_{\mathcal{M}_t} = m_k$ can be computed with (\ref{equation:forget}). The particle-based estimation of the marginal likelihood $p_k(\bm{y}_t|\bm{y}_{0:t-1})$ in (\ref{equation:likelihood}) is given by:
\begin{equation}
\label{equation:likelihood_pf}
p_k(\bm{y}_t|\bm{y}_{0:t-1}) \approx \sum^{N_s}_{i=1}\omega^i_{t-1} p_k(\bm{y}_t|\bm{x}^i_t).
\end{equation}

To mitigate the problem of particle degeneracy, resampling procedure is adopted to enhance the efficiency of particle evolution. Details of the particle filter are given in \cite{qi2019dynamic}.

\section{Closed-loop BMI Experiments}

\subsection{Ethics}
All clinical and experimental procedures in this study were approved by the Medical Ethics Committee of The Second Affiliated Hospital of Zhejiang University (Ethical review number 2019-158, approved on 05/22/2019) and were registered in the Chinese Clinical Trial Registry (Ref. ChiCTR2100050705). The informed consent was obtained both verbally from the participant and his immediate family members and signed by his legal representative.

\subsection{Participant}
The volunteer participant is a 74-year-old male who had a car crash and suffered from complete tetraplegia subsequent to a traumatic cervical spine injury at C4 level. The volunteer participant has the ability to move his body part above the neck and has normal linguistic competence and comprehension of all tasks. For the limb motor behavior, the patient scored 0/5 on his skeletal muscle strength and had a complete loss of limb motor control. 

\subsection{Microelectrode Array Implantation}

\begin{figure}[tb]
\centering
\includegraphics[scale=0.23]{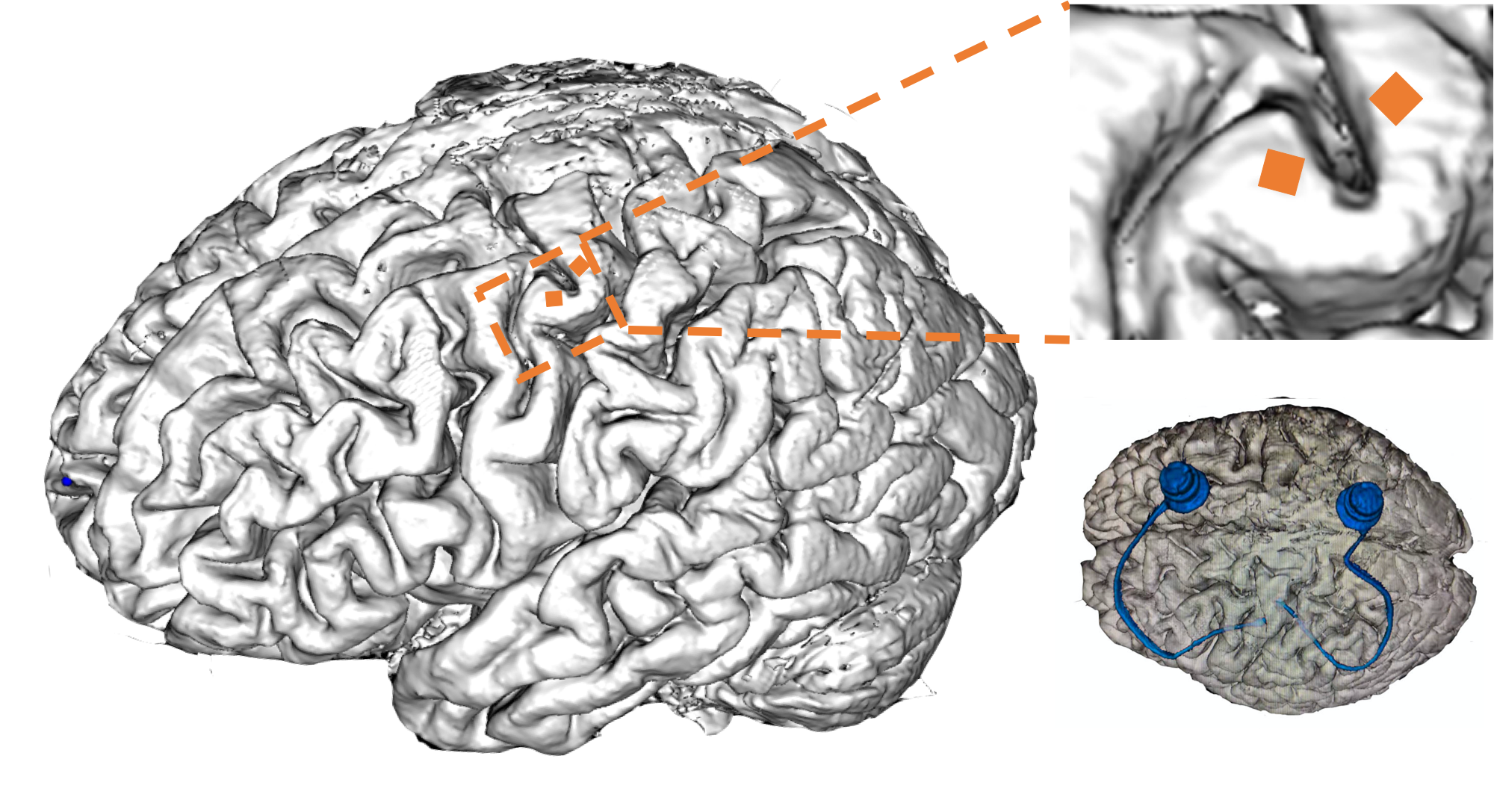}
\caption{{The implant locations of two Utah electrode arrays. The red squares illustrated the detailed implant locations on the primary motor cortex. The lower right image showed the positions of the array and the connectors (reconstructed with CT images).} 
}
\label{fig:locationArray}
\end{figure}

The participant was implanted with two 96-channel Utah intracortical microelectrode arrays (4mm $\times$ 4mm, Utah Array with 1.4 mm length, Blackrock Microsystems, Salt Lake City, UT, USA) in the left primary motor cortex, with one array in the middle of hand knob area (array-A) and the other located medially about 2mm apart (array-B). The implantation uses structural (CT) and functional imaging (fMRI) to guide the placement (Fig.~\ref{fig:locationArray}). The participant was asked to perform imagery movement of hand grasp and elbow flexion/extension with fMRI scanning to confirm the activation area of the motor cortex. The surgery used a robotic arm to hold the array hitter during the implantation. The participant was left for full recovery from the operation for a week before neural signal recording and BMI training tasks began.

\subsection{Neural Signal Recording}

The neural signal was recorded with the Neuroport system (NSP, Blackrock Microsystems). Neural activity was amplified, digitized, and recorded at 30 kHz. The thresholds for neural action potential detection were set at -6.5 to -4.5 times the root-mean-square of the high-pass filtered (250 Hz cut-off) full-bandwidth signal on each array separately with the Central software suite (Blackrock Microsystems), according to the condition of recorded signals. {For situations that neural signals interfered by external noises, we used lower thresholds to suppress the influence of noises. For each experiment day, the threshold was fixed.} Single- and multi-unit events were classified manually at the beginning of each daily task, which took about 25 to 35 min. The spike activity was converted to a firing rate in 20 ms bins and low-pass filtered using an exponential smoothing function with a 450 ms window. 

The participant performed BMI experiment tasks on workdays and rested on weekends. The BMI experiment took about 3 hours each day, including the preparation for signal recording, impedance testing, spike sorting, and main tasks. The experiment would be stopped once the participant reported being tired or in unusual physical conditions, such as having fever or urinary tract infection. 

\subsection{BMI Control Task}

The 2D cursor control is presented on a computer monitor placed 1.5 meters in front of the subject. In each trial, the participant was asked to control a blue cursor ball to reach a red target ball. In a successful trial, the distance between the centers of the blue and red balls would be less than a preset reach-threshold and held for no shorter than a pre-defined holding time. An auditory cue was given at the end of a trial to indicate whether it succeeded or failed. 
{In our experiments, there are a total of three BMI control tasks: radial-8 task with big balls, radial-8 task with small balls, and the random target pursuit (RTP) task with small balls}. The detailed settings of each task are given in Supplementary Table S1.

\begin{itemize}

\item
\textbf{Radial-8 / Big.} The task contains eight targets at the top, bottom, left, right, and the four corners of the screen. The targets display sequentially in clockwise order. In each trial, the subject moves the cursor to the target, and the cursor will automatically return to the center when the trial is completed or time out. 
{
\item
\textbf{Radial-8 / Small.} The settings are similar to Radial-8 / Big, only that the sizes of the cursor and target balls are relatively smaller, increasing the difficulty of the task.
\item
\textbf{RTP / Small.} In each trial, the target ball presents at a random position on the screen, and the subject controls the cursor to reach the target. There is no return stage in the RTP task, and the cursor directly begins from the previous target's position. 
}

\end{itemize}

\begin{figure}[tb]
\centering
\includegraphics[scale=0.31]{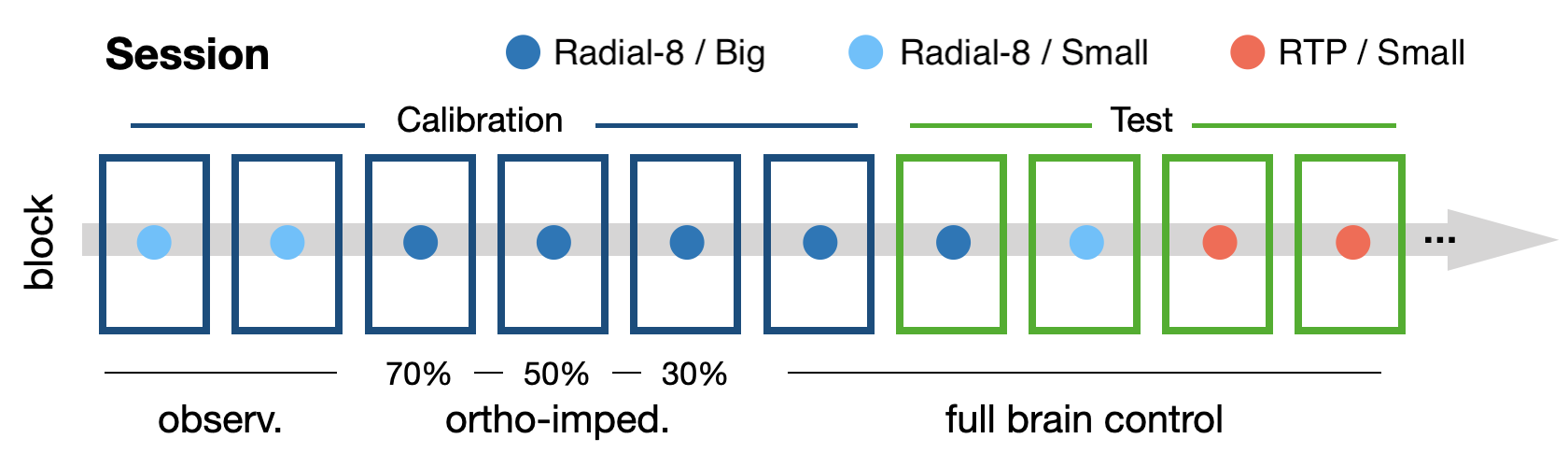}
\caption{{Description of the block design in each session.} Each session contains a calibration phase (with blue boxes) and a test phase (with green boxes). The calibration phase starts with two observation blocks and followed by three ortho-impedance assisted blocks and then a full brain control block. The test phase contains a Radial-8/Big block, a Radial-8/Small block, and several RTP blocks.
}
\label{fig:blocks}
\end{figure} 

\subsection{Closed-loop Decoder Calibration and Test} 

Each experiment session consists of two phases, calibration and test. Both phases contain multiple blocks, with 16 trials in each block. 
{
The calibration phase is similar to the methods used in \cite{collinger2013high-performance}.} It started with two observation blocks wherein the system performed the task automatically, with the participant closely observing the movement of the cursor. The neural signals recorded from the observation block were used to train the decoder initially. Ortho-impedance assistance was adopted in the following several blocks to further tune the decoder, but with a gradually decreased amount of assistance and an increased involvement of the participant ($0.7, 0.5, 0.3, 0$, see Supplementary Materials for details).

The last decoder in the training phase was used in the test phase, where no computer assistance was used. {The test phase begins at the 7$^{th}$ block, and the first two test blocks are the Radial-8 task with big and small balls respectively. The RTP task starts at the 9$^{th}$ block, and continues 3 to 10 blocks.} In the test phase, the timeout is set to 10 seconds (see Supplementary Materials for details).

\section{Results}

Experiments are carried out to evaluate the performance of DyEnsemble, including online BMI control, offline neural decoding, and simulations, where online comparison is mainly focused. In the online comparison, we evaluate the closed-loop BMI control performance of DyEnsemble in comparison with Kalman filter and Bayesian filter using single models. Then we analyze the dynamics of model weights under different conditions to demonstrate the effectiveness of dynamic ensemble. The results of simulation and offline decoding are presented in Supplementary Materials Section II-B and II-C.

\subsection{Online Decoding Performance}

Online experiments were carried out to evaluate the proposed DyEnsemble decoder in comparison with other approaches. An experiment day commonly contains two or three sessions according to the willingness of the participant. The experiments contain 12 days of data over about one month, and the experiments for each day are specified in Supplementary Table S3 and Table S4.

\subsubsection{Comparison with Kalman filter}

\begin{figure}[h]
\centering
\includegraphics[scale=0.37]{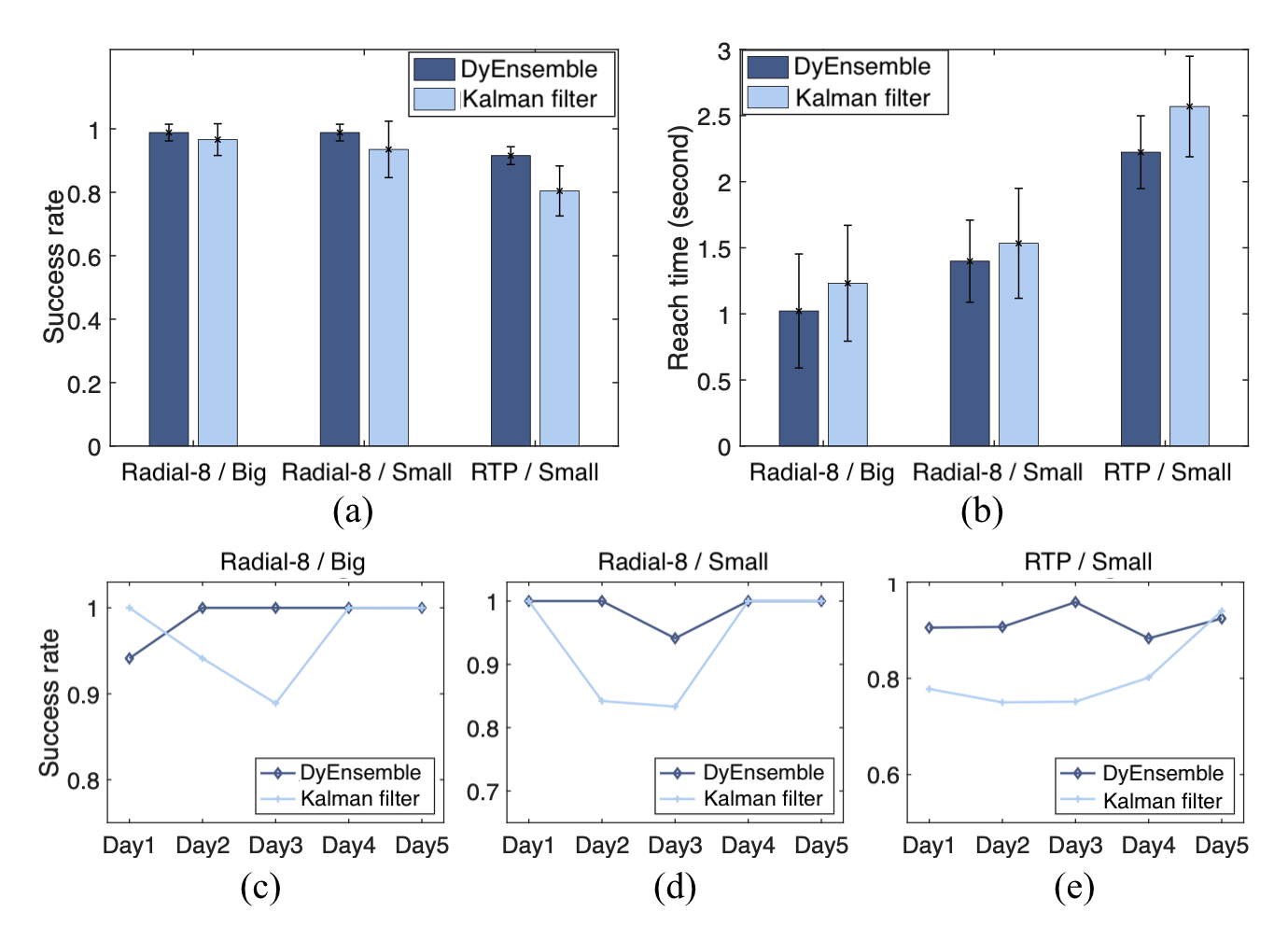}
\caption{{Online performance comparison of the DyEnsemble and the Kalman filter. (a) Comparison of success rate. (b) Comparison of reach time. (c) Comparison of success rate in different experiment days. } 
}
\label{fig:performance-compare}
\end{figure}

To compare the performance of DyEnsemble and Kalman filter, we evaluated the two decoders in two separate sessions at each day, and the session order was randomly assigned for fair comparison (see Table \ref{tab:compare}). 
Fig. \ref{fig:performance-compare} compares DyEnsemble and Kalman filter with success rate and reach time, in three tasks of Radial-8 with big and small targets, and the RTP task. As shown in Fig. \ref{fig:performance-compare} (a), DyEnsemble outperforms the Kalman filter with all three tasks, and the advantage is most significant in the RTP task, which is the most difficult. In the Radial-8 task, the average success rate of DyEnsemble is 98.8\% for both big and small targets, which is significantly higher than 96.6\% and 93.5\% of Kalman filter, respectively. 
With the RTP task, the performance of the Kalman filter drops to 80.4\%, while DyEnsemble keeps a high success rate of 91.6\%, which outperforms the Kalman filter by 13.9\%. Paired $t$-test is performed to evaluate the significance of the improvement. Results demonstrate that, compared with the Kalman filter, DyEnsemble improves the success rate significantly with  $p = 0.0089$ (paired $t$-test, a pair is defined as the success rate of DyEnsemble and Kalman filter of the same task in the same day). 
{ 
We further analyze the stability of online performance by the variance of the success rate over different days. Fig. \ref{fig:performance-compare} (c-e) compares the success rate of the three tasks across the five experiment days. With the DyEnsemble, the performance variances over different days are 6.92E-04, 6.92E-04 and 7.84E-04 for the three tasks respectively, which are 72\%-91\% lower than 2.51E-03, 7.90E-03 and 6.23E-03 of the Kalman filter. Results indicate that DyEnsemble performs more stably over days. 
}
DyEnsemble also reduces reach time with all three tasks. As shown in Fig. \ref{fig:performance-compare} (b), DyEnsemble decreases reach time by 17.0\%, 8.8\%, and 13.5\% for Radial-8 with big target, Radial-8 with small target, and the RTP task respectively. { However, the improvement of reach time is not significant compared with the Kalman filter (paired t-test, p=0.0560).
Therefore, results demonstrate that DyEnsemble improves accuracy (higher success rate),  stability (lower variance) and control efficiency (shorter reach time) in online BMI control. }

\begin{figure}[t]
\centering
\includegraphics[scale=0.35]{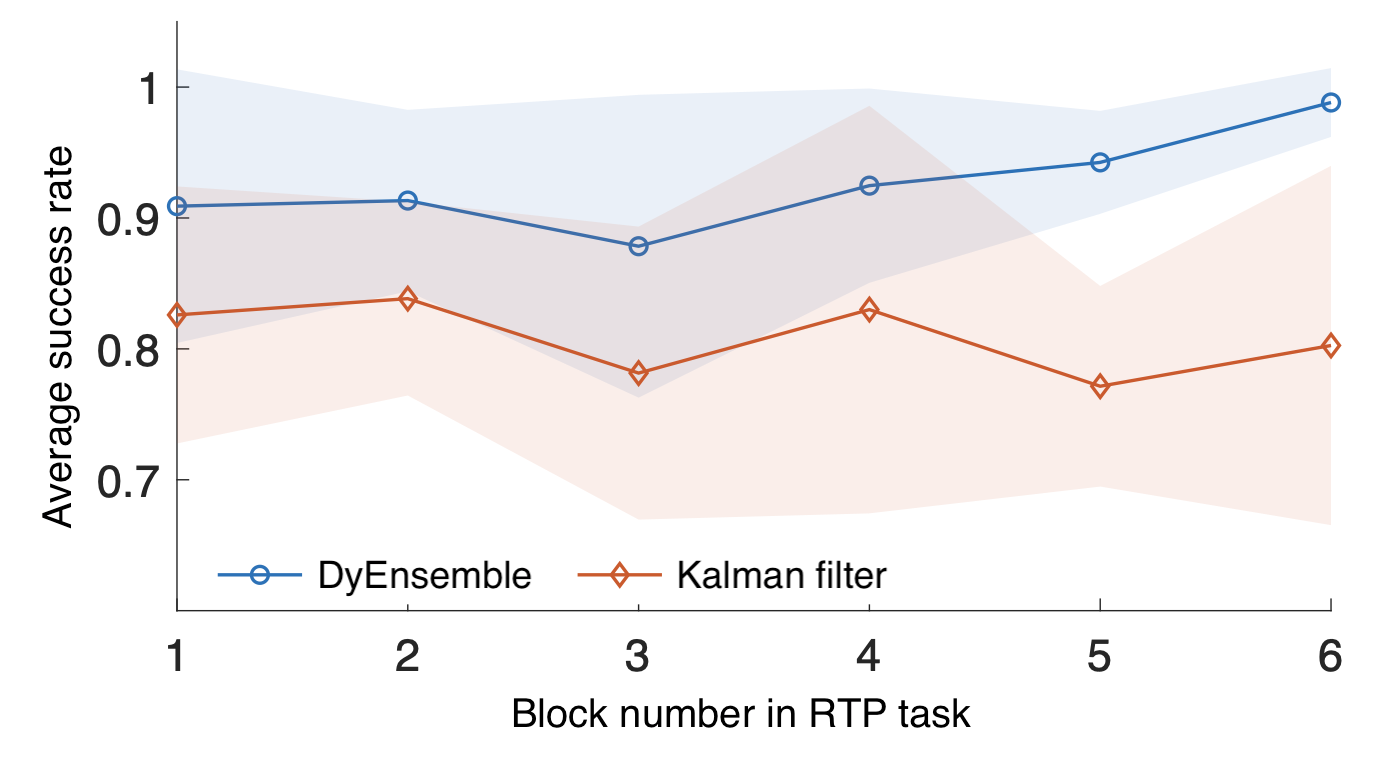}
\caption{{Success rate of BMI control in the RTP task. The x-axis is the block index starting with the first RTP block. } 
}
\label{fig:tracing-compare}
\end{figure}

{
In Fig. \ref{fig:tracing-compare}, we present the success rate over successive blocks for both DyEnsemble and Kalman filter in the RTP task. Results show that, with the DyEnsemble, the success rate increases rapidly along blocks, while the increase is not obvious with the Kalman filter. It suggests that the subject learns more efficiently with the DyEnsemble approach. While currently we only have online experiment results of one subject, more online evaluations will be carried out for further analysis. 
}

\begin{table*}[h]
\renewcommand{\arraystretch}{1.3}
\caption{Details of sessions on the experiment days.}
\label{tab:compare}
\centering
\begin{tabular}{c|ccccc}
\hline
Day / Date & Day1 / 12-08 & Day2 / 12-09 & Day3 / 12-14 & Day4 / 12-15 & Day5 / 12-16\\
\hline
Session 1 & Kalman filter (6-1-1-3)* & DyEnsemble (6-1-1-9) & DyEnsemble (6-1-1-10)& DyEnsemble (6-1-1-9)& Kalman filter (6-1-1-7)\\
Session 2 & DyEnsemble (6-1-1-5)& Kalman filter (6-1-1-4) & Kalman filter (6-1-1-8)& Kalman filter (6-1-1-7)& DyEnsemble (6-1-1-7)\\
\hline
\end{tabular}
\begin{flushleft}
{
* The numbers in the parentheses indicate the number of blocks for each task. The numbers are in the form of (T-B-S-R), where T,B,S, and R are the block numbers for training, test with radial-8-big, test with radial-8-small, test with random target pursuit task, respectively.}
\end{flushleft}

\end{table*}

\subsubsection{Comparison with single models}

\begin{figure}[h]
\centering
\includegraphics[scale=0.3]{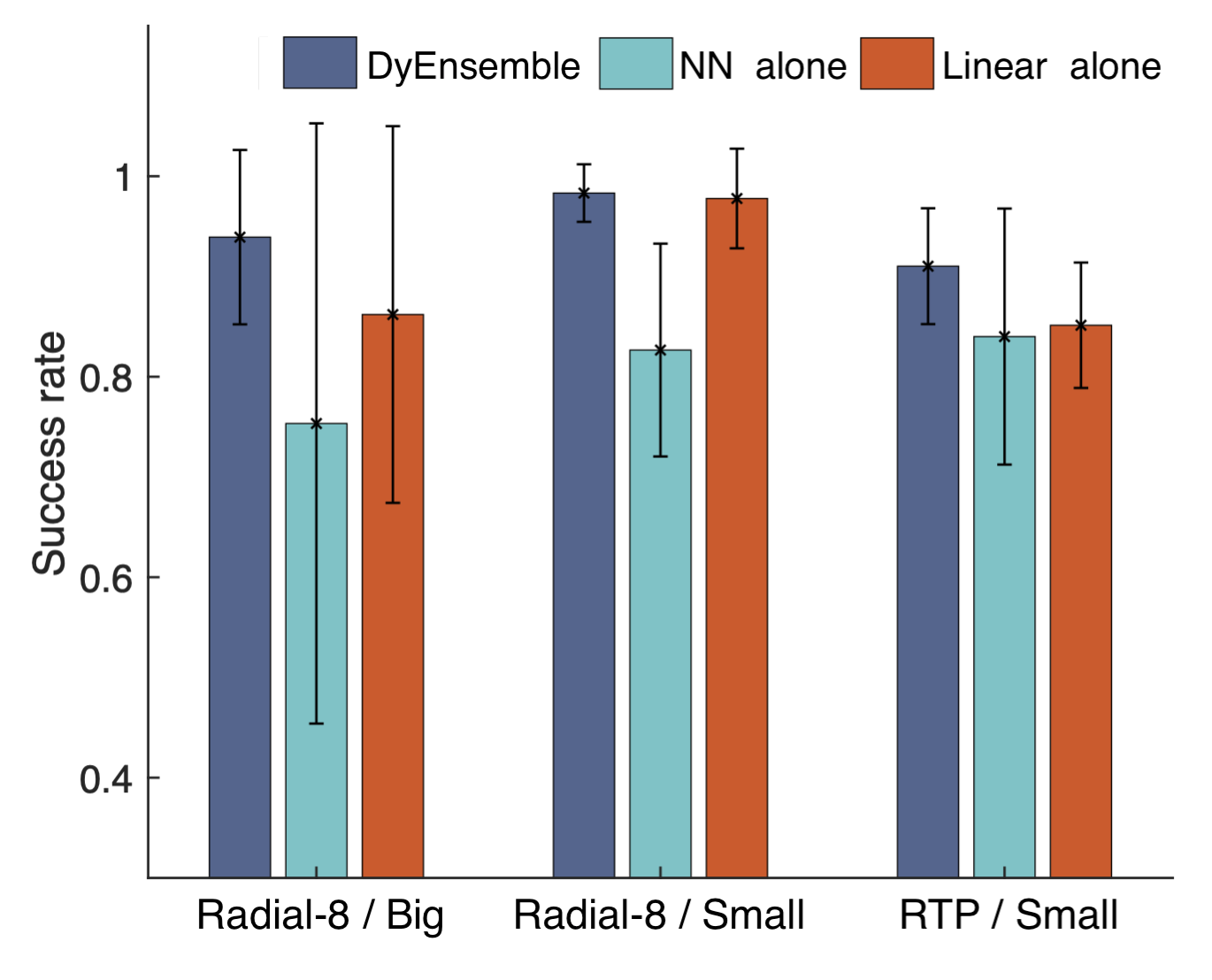}
\caption{{Performance comparison between the multiple-model DyEnsemble and DyEnsemble using the single models alone.} 
}
\label{fig:single_model}
\end{figure}

Online experiments were carried out to compare the DyEnsemble with single models using a linear model or a neural network (NN-2) only. For single models, the particle filter is applied similar to DyEnsemble for a fair comparison. The details of the online experiments are given in Table S3. The results are illustrated in Fig. \ref{fig:single_model}. 
Overall, DyEnsemble demonstrates the highest success rate and stability with all three tasks of Radial-8 with big and small targets, and the RTP task. The success rates of DyEnsemble are 93.9\%, 98.3\%, and 91.0\%, respectively, for the three tasks. The success rates of single linear and network models are 86.2\%, 97.8\%, 85.1\%, and 75.3\%, 82.7\%, 84.0\%, for the three tasks, respectively. { DyEnsemble significantly outperformed both the single linear model and the single NN model, with the $p$-values of $t$-test of 0.0465 and 0.0009, respectively. }
In the experiment, we observed that using a single model gave an unstable performance. For several times, the training process has to be aborted due to bad control performance (one time for the linear model and two times for the NN model, as shown in Supplementary Table S4). Using a single neural network obtained the worst performance. The results demonstrate the necessity and effectiveness of dynamic ensemble.

\subsection{Model Adaptation in Online Control} 

In online BMI control, we observed that with the Kalman filter, the participant occasionally had trouble in moving the cursor to certain directions (such as the upper right), which led to poor control performance. While DyEnsemble performs more robust and stable compared with the Kalman filter. Analysis was carried out to investigate how the model ensemble adapts to neural signals. Firstly, we tracked the model weight changes in single trials to demonstrate the effectiveness of dynamic ensemble. Then we analyzed the dynamic changes of model weights under different cursor speeds. 

\subsubsection{Analysis of the dynamic ensemble process} 

\begin{figure*}[htb]
\centering
\includegraphics[scale=0.47]{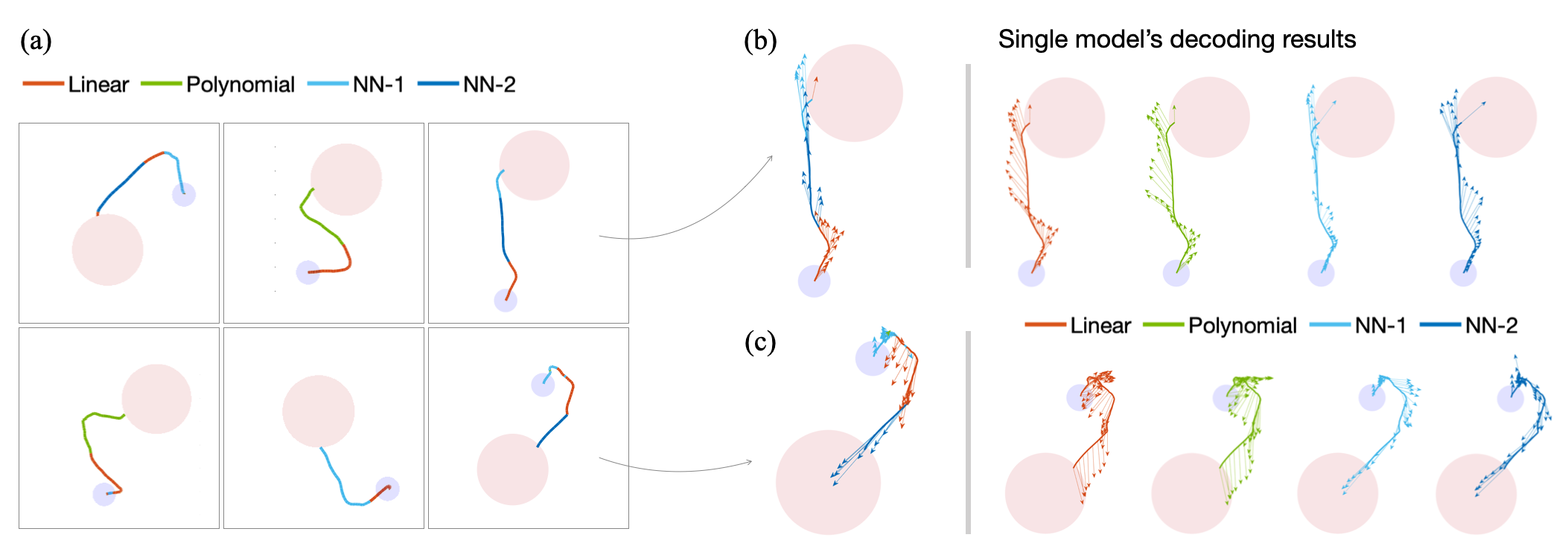}
\caption{{Online performance comparison of DyEnsemble and Kalman filter. {(a) illustrates examples of the RTP task trials which contains dominant model switches. (b) and (c) present two specific RTP trials, where the velocities predicted with the DyEnsemble (on the left) and the single models (on the right) are illustrated to demonstrate the effectiveness of dynamic ensemble.} DyEnsemble can switch to proper models at each time slot such that more accurate and robust performance can be achieved. } 
}
\label{fig:model-switch-big}
\end{figure*}

Here we investigate the dynamic model ensemble process to demonstrate how DyEnsemble adaptively adjusts itself with changes in neural signals. We visualized the model weight change process in single trials. In Fig. \ref{fig:model-switch-big} (a), we especially illustrate several RTP task trials with obvious model switching processes (which are mostly curvy). In each subfigure in Fig. \ref{fig:model-switch-big} (a), the color of trajectory segments indicates which model was assigned with the largest weight. We can see that the dominant models altered during a single RTP task trial process.

In Fig. \ref{fig:model-switch-big} (b) and (c), we further analyze what causes the model weights to change with two examples. We illustrate the decoded velocity at each time step with an arrow for both the ensemble model (on the left), and single models in the model pool (on the right). 
With the RTP task trial in Fig. \ref{fig:model-switch-big} (b), the initially dominant model was linear (stage I), and then switched to NN-2 (stage II). From the right panel, we find that at stage II, the speed direction decoded by the linear model deviated from the target, which could lead to inaccurate estimations. While with the NN-2 model, the decoded speed direction was pointed to the target. At this point, the DyEnsemble algorithm dynamically switched the dominant model from linear to NN-2. Thus the cursor could be correctly controlled toward the target. While with the linear model alone, the cursor could be deviated to the left and led to a failure trial. When the cursor was close to the target, the estimation of NN-2 deviated, and the DyEnsemble model adaptively switched to NN-1, which was more accurate. Fig. \ref{fig:model-switch-big} (c) shows the model switching process with another RTP task trial. Similar results can be found that the DyEnsemble model switched its dominant models adaptively to make more accurate online control.

The results demonstrate that, with the variability in neural signals, different models can show varying performance in an online control process. The variation and unstable decoding performance of single models suggests the necessity and importance of dynamic model ensemble for robust BMI control. Results demonstrate that the DyEnsemble model can adaptively adjust the ensemble of models to cope with variability in neural signals, thus effectively improving the intention estimation accuracy and stability in online BMI control.

\subsubsection{Model adaptation to the cursor speed} 

\begin{figure}[t]
\centering
\includegraphics[scale=0.3]{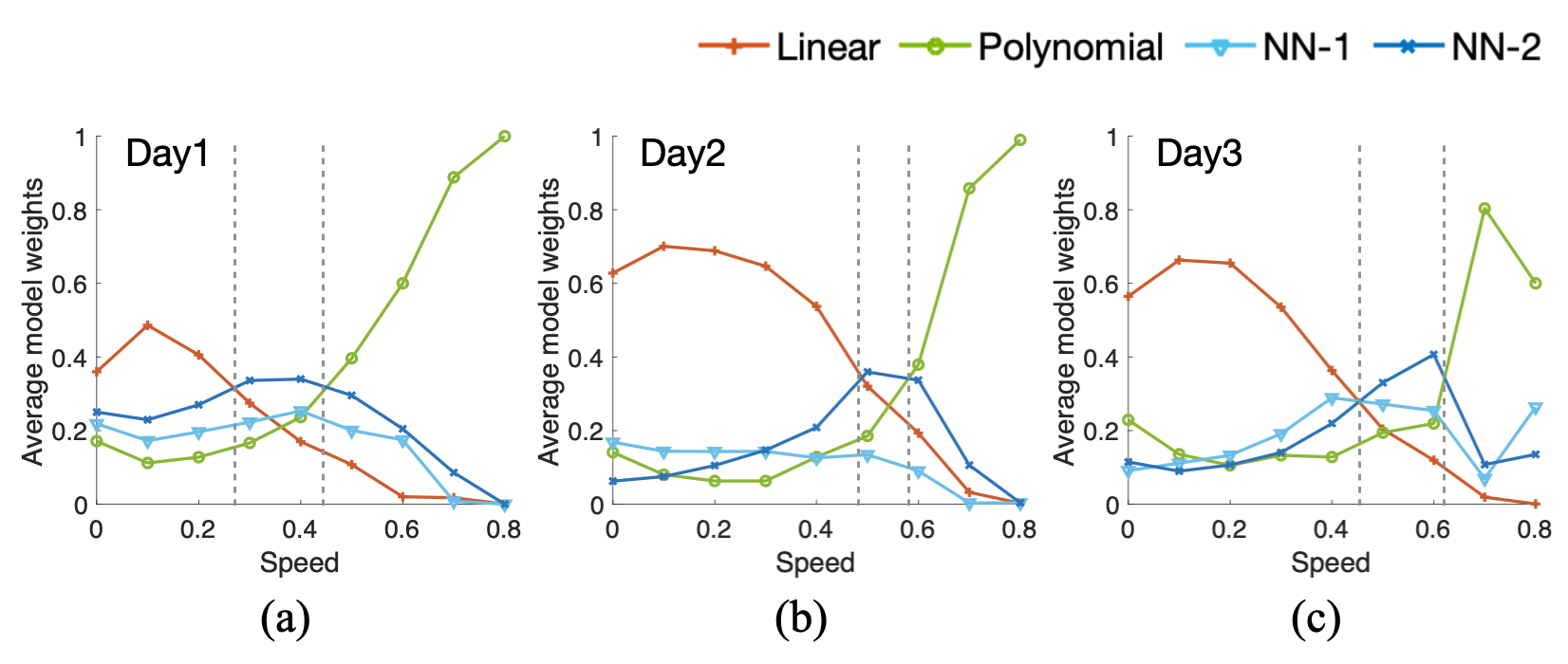}
\caption{{Changes of model weights with the cursor speed.} 
}
\label{fig:velocity}
\end{figure}

Studies have reported that neural activities change with cursor speed during BMI control \cite{sussillo2016making, churchland2006preparatory}. Here we analyze the dynamic ensemble by examining model weight distributions across different speed conditions. Fig. \ref{fig:velocity} illustrates the results of three DyEnsemble sessions. It is interesting to find that DyEnsemble's preference of models is different with different speed ranges. Specifically, the speed can be roughly divided into three ranges, including low, medium, and high, and the three ranges are alternately dominated by linear, neural networks, and polynomial models, respectively. We also observed that the pattern was mostly consistent among different sessions, only that the division of speed ranges shifted among sessions.

In Fig. \ref{fig:velocity} (b), the peak of the linear model is around 0.1, and the weight drops quickly at the high-speed range, and the weight is almost zero when the speed is higher than 0.7. It indicates the limitation of the linear models that they are only accurate in a certain speed range. While in online control, the cursor speed is mostly in the low-speed rate (69\%, 75\%, and 71\% for the three sessions, respectively). Thus linear decoders usually present good performance. Contrary to the linear model, the polynomial model demonstrates high weights at the high-speed side, and the weight increases rapidly when speed is over 0.5. The dominant weight of polynomial models at high speed may indicate that there exists a speed-related higher-order variable in neural encoding of motor behaviors. The middle-speed range is covered by the NN-2 model. Similar results can be found in both Fig. \ref{fig:velocity} (a) and (c). 

{
The results show the dynamic complementary property of different models, which also helps explain how the dynamic ensemble model works. 
As neural encoding function can occasionally shift, different models have diverse abilities in describing the neural encoding functions in each time slot (as shown in Fig. \ref{fig:model-switch-big} and Fig. \ref{fig:velocity}). In DyEnsemble, the dynamic re-assembling process can assign proper models with higher weights in each time slot, thus improving handling with variability. 
}

\subsection{Model Adaptation in Closed-loop Calibration} 

Here we analyze the model weight changes in the closed-loop BMI calibration. During calibration, we guided the subject to learn brain control by gradually decreasing the computer assistance and ortho-impedance assistance percentage. As the participant of brain control increases, the subject adjusted the neural signals as the effect of learning. Adaptively, the DyEnsemble model recalibrated the models in the pool and adjusted the model ensemble to cope with changes in neural signals. 

\begin{figure}[t]
\centering
\includegraphics[scale=0.38]{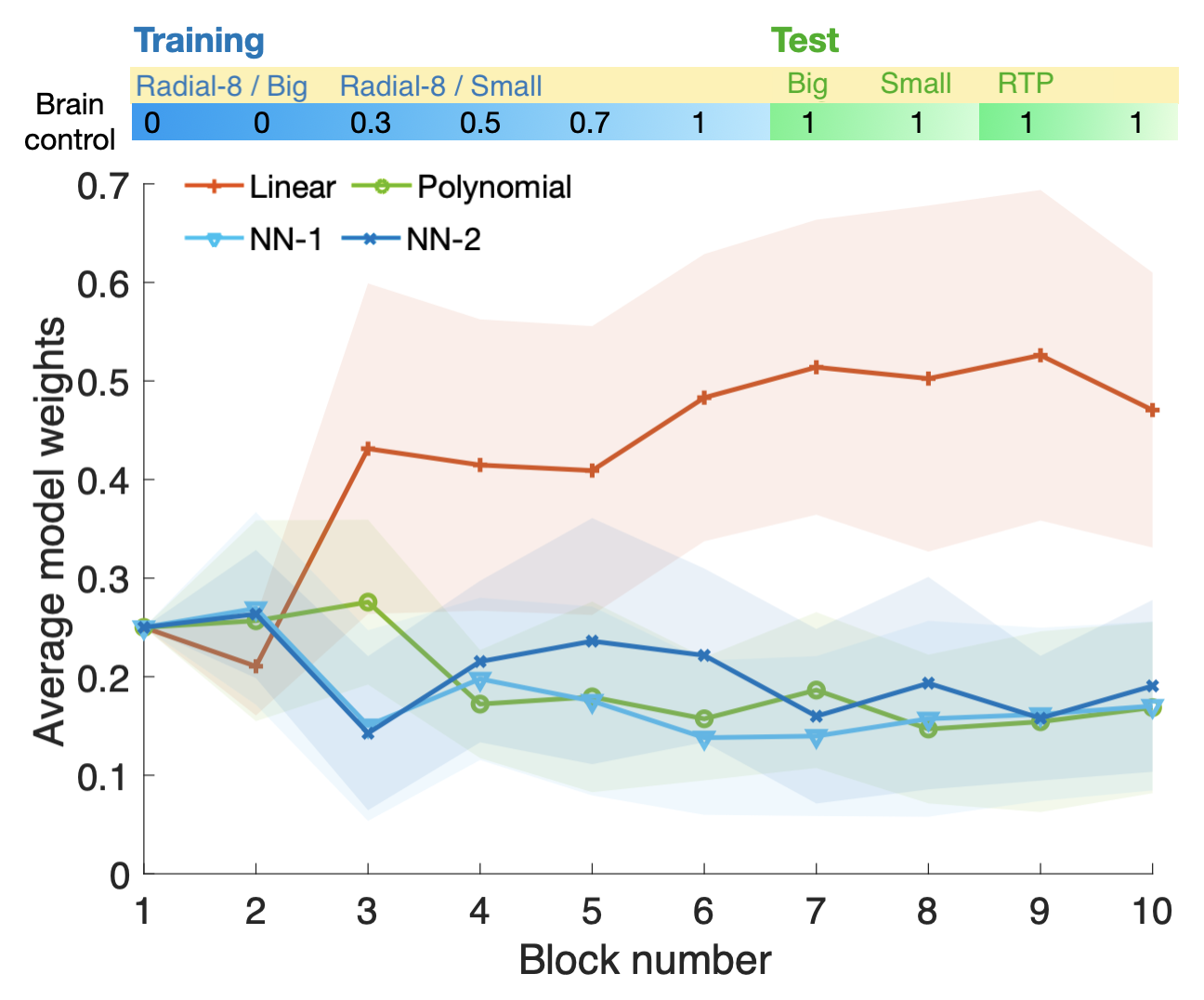}
\caption{{Changes of model weights with different blocks. } 
}
\label{fig:weight-curve}
\end{figure}

In Fig. \ref{fig:weight-curve}, we illustrate the weights of the four models in DyEnsemble across blocks. For each model, we average the weights over blocks. All the DyEnsemble sessions (a total of 11 sessions) across experiment days are employed. We observe that nonlinear models obtain higher weights in the observation blocks (block 2), while as the brain control percentage increases, the weight of the linear model increases and becomes dominant after the third block. During the calibration phase (block 1-6), the weight of the linear model increases continuously and becomes stable in the test phase. The pattern of weight change is consistent over different sessions and days (see Supplementary Materials Fig. S1). 
{The results demonstrate that, as the subject gradually controls the cursor, the model ensemble of DyEnsemble dynamically changes to cope with the changes in neural signals. 

It is interesting to find that, when the subject fully controls the cursor, the model ensemble is a mixture of linear and nonlinear models, where the linear model takes the dominant  part. We suppose that the performance improvement is brought by the brief switches of models (as shown in Fig. \ref{fig:model-switch-big} and Fig. \ref{fig:velocity}), especially at high speeds that are poorly captured by linear models.

}

\section{Discussion}

The variability of neural signals has been a major challenge for robust and reliable motor BMI systems. Most existing neural decoding studies focus on more accurate modeling of functional relationships between neural signals and motor kinematics, while how to maintain a high performance given the variability of neural signals is not well addressed. 
The proposed DyEnsemble framework provides a novel perspective of thinking for adaptive online decoding in BMI systems and demonstrates its potential for robust and reliable BMI control. 

\subsection{Comparison with Existing Adaptive Decoders }

Existing studies with adaptive BMI decoders can be divided into two categories: recalibration and dynamic modeling. 
{
Recalibration approaches cope with variability in neural signals by periodically recording both kinematic and neural signals and retrain decoders \cite{gilja2015clinical, shanechi2016robust, brandman2018rapid, wen2021rapid}, and the state-of-the-art approaches have enabled rapid recalibration within several minutes \cite{brandman2018rapid, wen2021rapid}. 
However, commonly the recalibration process would interrupt the BMI use process, which hinders the user experience. 
}

Compared with recalibration-based decoders, the proposed DyEnsemble is able to tune the decoding model to cope with changes in signals without task or target information.
Dynamic modeling approaches attempt to build a model to describe the dynamical changes of neural activities \cite{eden2004dynamic, wang2008tracking, wang2015tracking}. 
Study \cite{liyanage2013dynamically} proposed a Bayesian ensemble classifier, which demonstrated superior performance in non-stationary EEG signal processing. Compared with \cite{liyanage2013dynamically}, DyEnsemble accounts for the sequential property of neural variability by extending the Bayesian ensemble to a state-space version. 
Other studies from this aspect include dual Kalman filter \cite{wang2008tracking} and adaptive point-process \cite{wang2016tracking}. However, they usually rely on strict assumptions of neural variability, and the performance can be degraded when the assumptions are not conformed. In DyEnsemble, the dynamic model adjustment process is completely data-driven and does not have strong assumptions in neural signals.

From the Bayesian perspective, the proposed DyEnsemble approach extends existing Bayesian decoders such as Kalman filters by accounting for model uncertainty in the neural decoding tasks, which can alleviate the risks brought by the overconfident inference with a fixed model \cite{fragoso2018bayesian, montgomery2010bayesian}, thus improving the robustness of neural decoders.

\subsection{DyEnsemble Provides a Novel Framework for Robust Neural Decoding}

The presence of neural variability can lead to unstable performance, which has been a challenging problem for BMI systems. The variability in neural activities occurs continuously in association with the dynamics of neural systems, instabilities in signal recordings, or caused by different brain states and conditions, and can be even more complicated with the influence of closed-loop feedbacks online \cite{sussillo2016making, suway2018temporally, even2017augmenting, degenhart2020stabilization}. 
In this study, with an DyEnsemble model using a model pool constituted of several linear and nonlinear models, we mainly focus on the variability of neural encoding in closed-loop BMI control. 

{ 
More broadly, DyEnsemble frames a novel and flexible framework for robust neural decoding. From this aspect, this study can be considered as an example of DyEnsemble, which uses diverse function forms to deal with variability of neural encoding functions.
Our previous study \cite{qi2019dynamic} is another example of DyEnsemble. In \cite{qi2019dynamic}, we proposed to build a linear model pool, where the model candidates were generated with linear models by random channel dropout and weight perturbation to cope with nonstationary noises in channels. }By designing and selecting different model pools with specific purposes, it is feasible to cope with different types of variability. 

One limitation of DyEnsemble lies in that the performance highly depends on the construction of the model pool. For an effective model ensemble, the models in the pool should be diverse in the decoding ability and cover different variability in neural signals. In future work, different ways of building proper model pools to deal with diverse variability are to be studied.

\subsection{DyEnsemble is a Flexible Encoder-based Decoding Model}

Traditional Bayesian filters such as the Kalman filter can be considered as an encoder-based neural decoder, while only one fixed encoding model is applied. DyEnsemble extends Bayesian filters to a more flexible encoder-based decoder. In DyEnsemble, the model pool is equipped with multiple encoding models, and the dynamic ensemble process dynamically weights the encoders by how much they can explain the observed neural signals. With the DyEnsemble framework, neural decoders can be more flexibly and friendly constructed by designing a pool of possible encoders, and then letting DyEnsemble decide which model to use according to neural signals. 

Encoder-based neural decoders have special advantages, because they are more interpretable compared with decoders that model the opposite causality from neural activity to kinematic variables. 
Neural encoding models, from classical tuning curves \cite{georgopoulos1982relations} to recently developed nonlinear functions \cite{liang2019deep} contain rich information for neural decoding from the perspective of neuroscience. DyEnsemble can use encoding models directly for decoding, which can benefit from the development of encoding models, and provides a framework for evaluation and comparison of encoding models.

\section{Conclusion}
Toward robust motor BMI systems against neural variability, we propose a novel adaptive  neural decoding approach, namely the DyEnsemble, which is able to adapt to nonstationary neural variability by dynamically adjusting the decoder according to neural activities. Different from existing decoders which rely on one fixed model, DyEnsemble learns a pool of models and dynamically assembles them along with changes in neural signals. 
Results demonstrated that the model ensemble of DyEnsemble can adaptively change to cope with variability in neural encoding, thus more accurate and robust performance can be achieved. DyEnsemble provides a flexible adaptive neural decoding framework, and by designing and selecting different model pools with specific purposes it is feasible to cope with different types of variability.

\section{Acknowledgment}

The authors would like to thank Yangang Li and Zijun Wan for the support in the online experiments, and thank Huaqin Sun for the help with the simulation experiments.
This work was partly supported by the grants from National Key R\&D Program of China (2018YFA0701400), Natural Science Foundation of China (61906166, 61925603), Zhejiang Lab (2019KE0AD01), Key R\&D Program of Zhejiang (2022C03011), the Project for Hangzhou Medical Disciplines of Excellence, the Key Project for Hangzhou Medical Disciplines, the Chuanqi Research and Development Center of Zhejiang University, and the Fundamental Research Funds for the Central Universities, the Starry Night Science Fund of Zhejiang University Shanghai Institute for Advanced Study (SN-ZJU-SIAS-002).

\ifCLASSOPTIONcaptionsoff
  \newpage
\fi

\bibliographystyle{IEEEtran}

\bibliography{tbme}

\end{document}


\title{Supplementary Materials for \\
Adaptive Ensemble Bayesian Filter for Robust Control of a Human Brain-machine Interface }


\maketitle

\IEEEpeerreviewmaketitle

\section{Methods}

{\color{red}

\subsection{System Configuration Details}

A calibration phase was adopted in each session, taking about 10 minutes. The calibration phase consists of two blocks for observation, and four blocks with assistant-aided control, with decreasing ortho-impedance assistant ratios of $0.7, 0.5, 0.3, 0$. In the calibration phase, the decoder was recalibrated after each block using data from all the preceding blocks. The last decoder in the training phase was used in the test phase.

\subsubsection{Ortho-impedance}

The ortho-impedance reserves the projection of the control velocity in the ideal direction, i.e., the direction from the cursor to the target, and decreases the projection of control velocity in perpendicular to the ideal direction, preventing deviation from the target direction. 

\subsubsection{Decoded, planner and control velocity}
There are three types of velocity in the closed-loop calibration and test: decoded velocity, control velocity and planner velocity. 
\begin{itemize}
\item \textbf{Decoded velocity.} The decoded velocity is the velocity predicted from the neural signals using the neural decoders.
\item \textbf{Planner velocity.} The planner velocity is the ideal velocity computed given the source and target positions. The direction of the planner velocity directly points to the target. The speed is set by considering both the acceleration and the maximum speed. If the time required for reducing the current speed to zero is less than the time required for using the current speed to reach the target, in other words, if the current cursor position is close to the target, the cursor will decelerate. Otherwise, the cursor will accelerate.
\item \textbf{Control velocity.} The control velocity is the velocity used to control the cursor in the screen. In the observation blocks, control velocity is equal to the planner velocity. In the training blocks, control velocity is the decoded velocity with the ortho-impedance assistant. In the test blocks, control velocity is equal to the decoded velocity.
\end{itemize}

\subsubsection{Data used for calibration}
For each trial, there was a timeout of 3 seconds, and only trials completed within the timeout were considered a success. For each target, the maximum number of attempts is 3. Only the successful trials are used for training. 

In the calibration of decoders, only neural data in the reach stage were adopted, along with the corresponding planner velocity sequence. After each block, the decoder is recalibrated using data from all the preceding blocks.

\subsubsection{Settings of the tasks}

\modify{The parameters of both Radial-8 and the RTP tasks are configurable in the task settings, including the distance of the target from the screen center, the diameter of the target ball, the maximum time for a trial, the reach threshold, and the minimum holding time.} The detailed settings of the three tasks are shown in Table \ref{tab:task}.

\begin{table*}[h]
\renewcommand{\arraystretch}{1.3}
\caption{Configurations of the cursor control task.}
\label{tab:task}
\centering

\begin{tabular}{c|ccccccc}
\hline
Task  & Cursor size & Target ball size & Reach-thresh & Holding time \\
\hline
{Radial-8 / Big} & 0.07&0.13 & 0.1 & 0.05\\
{Radial-8 / Small} & 0.05&0.1 & 0.075& 0.05 \\
{RTP / Small} & 0.05&0.1 & 0.075 & 0.05\\
\hline
\end{tabular}
\end{table*}

\subsubsection{Computation of cursor position}

The cursor position displayed on the screen is obtained by integrating the control velocity:
\begin{equation}
\bm{p}_{t} =  \bm{p}_{t-1} + \bm{v}_{t} \cdot {Dt}
\end{equation}
where $\bm{p}_{t}$ is the cursor position at time t, $\bm{v}_{t}$ is the control velocity of the cursor, and $Dt$ is the time bin of 20ms.

\subsection{Details for the Kalman Filter}
In the online comparison, the Kalman filter is a velocity Kalman filter with a linear-Gaussian state-space model:
        \begin{align}
        \label{equation:state}
        \bm{x}_t=A\bm{x}_{t-1}+\bm{b}+\bm{p}_{t-1}\\
        \label{equation:observation}
         \bm{y}_t = H\bm{x}_t+\bm{q}_{t-1}
         \end{align}
where (\ref{equation:state}) and (\ref{equation:observation}) are the state transition and measurement model, respectively. In the equations, $t$ denotes the time step. $\bm{x}_t$ is a $2 \times 1$ vector of $[v_x,v_y]$. $\bm{y}_t$ is the firing rate of neurons in a 20 ms bin at time $t$, which is a $C \times 1$ vector with $C$ denoting the number of neurons. $A$ and $\bm{b}$ are the transition matrix and bias of the transition model, with $\bm{p}_{t} \sim N(0, \sigma^2_{p})$ is the i.i.d Gaussian state transition noise. $H$ is the measurement function and $\bm{q}_{t} \sim N(0, \sigma^2_{q})$ is the i.i.d Gaussian measurement noise.

The Kalman filter is calibrated in a same closed-loop paradigm as AdaEnsemble. The parameters of $A$, $\bm{b}$ and $H$ are estimated with the least square algorithm, using the neural signals and the corresponding planner velocity. $\bm{p}_{t}$ and $\bm{q}_{t}$ are estimated with the fitting residuals.
}

\section{Results}

\subsection{Model Adaptation in Closed-Loop Calibration}

\begin{figure*}[t]
\centering
\includegraphics[scale=0.49]{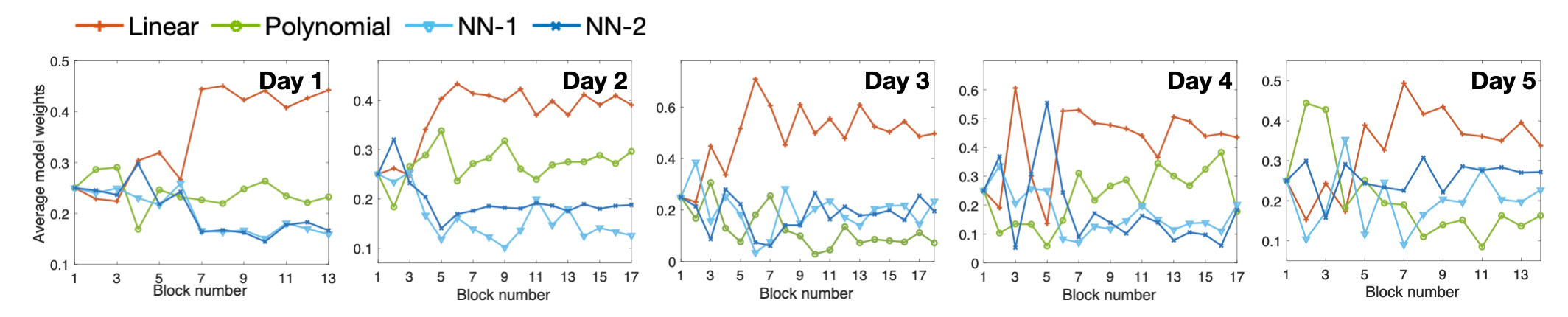}
\caption{Evolution of model weights in AdaEnsemble in first five experiment days.} 
\label{fig:curve}
\end{figure*}

Here we analyze the model weight changes during the closed-loop BCI calibration. In Fig. \ref{fig:curve}, we illustrate the model weight evolution in AdaEnsemble in the first five experiment days. Results show that,  consistent patterns are observed over different sessions and days. 

\subsection{Simulation Results}

\begin{figure}[t]
\centering
\includegraphics[scale=0.45]{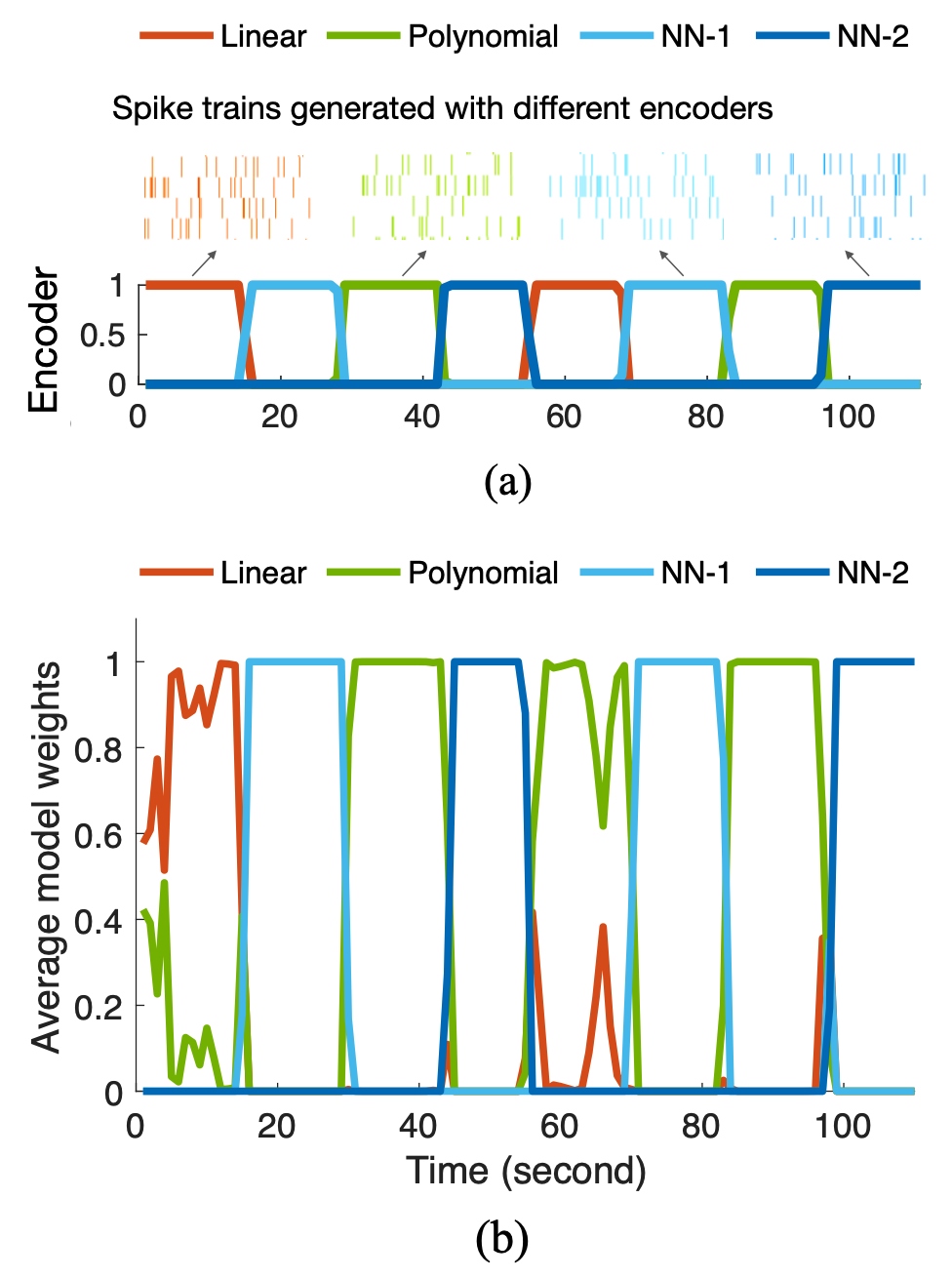}
\caption{Simulation of encoder change and the model dynamic estimation with AdaEnsemble. (a) Simulation of neural data with changing encoders. (b) Model weight estimated with AdaEnsemble approach.} 
\label{fig:simulation}
\end{figure}

To evaluate whether the proposed AdaEnsemble approach can effectively adjust the model ensemble along with changes in neural signals, simulation experiments are carried out. To simulate controllable neural activity changes, we adopt a data-driven encoder learning approach in study  \cite{liang2019deep}. Specifically, we employ the ZJUNDD neural signal dataset of monkey center-out task \cite{zhou2014decoding}, and learn several encoders with different encoding models, including linear, polynomial, and neural networks. Then, we generate a neural signal sequence with changing variability by sequentially altering neural encoders. Fig. \ref{fig:simulation} (a) illustrates the neural signals generated with changing neural encoders. 

In the AdaEnsemble decoding process, we directly adopt the encoders used for data generation as the model pool, and examine the dynamic model assemble process. In Fig. \ref{fig:simulation} (b), we illustrate the model weights estimated by AdaEnsemble. The forgetting factor is set to 0.98, and the model weights are smoothed in time for illustration convenience. Overall, the AdaEnsemble approach is able to dynamically re-assemble the models to cope with changes in neural signals. In Fig. \ref{fig:simulation} (b), the dominant model of AdaEnsemble alters in time along with the changes of encoders, and the sequence is mostly consistent with the ground truth. Specifically, the initial encoder in the signal generation is the linear encoder, and the linear model obtains the dominant weight in decoding. At about the 15$^{th}$ second, the encoder gradually changes to the NN-1 model, and accordingly, the weight of NN-1 increases quickly to almost 1 in AdaEnsemble. The only incorrect ensemble appears at about 57$^{th}$-70$^{th}$ seconds, where the encoder is linear, while AdaEnsemble mainly uses the polynomial model. It may be because the polynomial model covers most information of the linear model in this neural dataset. Results demonstrate that, the proposed AdaEnsemble approach can correctly adjust the model ensemble along with changes in neural signals.

We further evaluate the decoding performance of AdaEnsemble in comparison with single models,  static model (Kalman filter) and fixed ensemble (BMA) in Table. \ref{tab:simulation_results} with three simulation datasets. The datasets are simulated with four encoders as in Fig. \ref{fig:simulation} (a), only that the original neural data are the dataset1, dataset2, and dataset3 in \cite{zhou2014decoding}, respectively. Overall, AdaEnsemble models obtain the best decoding performance with the highest CC and lowest MSE for all three datasets. With dataset1, AdaEnsemble achieves CCs of 0.894, 0.898, and 0.892 with forgetting factors of 0.98, 0.5, and 0.1, respectively. With single models, the decoding CCs are lower. NN-1 obtains the best performance of 0.862 in the single model group, followed by NN-2 (0.833) and polynomial (0.830) models, and the linear model obtains the lowest CC of 0.813. The results demonstrate the necessity of multiple models with changing neural signals. To demonstrate the advantages of the adaptive ensemble, we compare AdaEnsemble with Bayesian model averaging (BMA), which uses fixed model weights. Results show that BMA achieves CCs of 0.876, 0.912, and 0.900 for the three datasets respectively, which outperforms single model-based decoders, while the performance is inferior to AdaEnsemble. With the dynamical model ensemble, AdaEnsemble further improves the CCs by about 2.5\%. Results demonstrate the effectiveness of the dynamic ensemble process.

\begin{table}[tb]
\renewcommand{\arraystretch}{1.3}
\caption{Configurations of the cursor control task.}
\label{tab:simulation_results}
\centering
\begin{tabular}{c|ccccccc}
\hline
\multirow{2}{*}{Model}  & \multicolumn{2}{c}{Dataset1} & \multicolumn{2}{c}{Dataset2} &\multicolumn{2}{c}{Dataset3} \\
 \cline{2-7} & CC & MSE & CC & MSE& CC & MSE\\
\hline
AdaEn (0.98) & 0.894  & \textbf{0.014} & 0.927 & \textbf{0.010}&\textbf{0.909}  &\textbf{0.012} \\
AdaEn (0.5) & \textbf{0.898}  & \textbf{0.014}   & \textbf{0.929} & \textbf{0.010}&0.906  &0.013   \\
AdaEn (0.1)&  0.892  & 0.015 & 0.925 & 0.011&0.904  &0.013 \\
\hline
Linear&  0.813  & 0.024&0.877 & 0.016 & 0.870  &0.017  \\
Polynomial&   0.830  & 0.022 &0.885 & 0.015 &  0.874  & 0.017 \\
NN-1&   0.862  & 0.018 &0.894 & 0.015 &  0.881  & 0.016 \\
NN-2&   0.833  &0.022 &0.862  & 0.019 &  0.835  & 0.021 \\
\hline
Kalman filter &   0.873&0.017&0.894& 0.014&  0.888& 0.015\\
BMA&  0.876  & 0.017 & 0.912& 0.013&0.900&0.015\\
\hline 
\end{tabular}
\end{table}

\begin{table*}[h]
\renewcommand{\arraystretch}{1.3}
\caption{Data details of session in comparison with Kalman filter.}
\label{tab:compare_1}
\centering
\begin{tabular}{c|c|c|c|c|c|c|c}
\hline
Date & Session &Block& Decoder & Train Acc & Radial-8 / Big Acc & Radial-8 / Small Acc & \modify{RTP / Small Acc}\\
\hline
2020-12-08 & 1 & 11 (6-1-1-3)* & Kalman & 97.96 \% & 100 \%  & 100 \% & 77.78 \% \\
\hline
2020-12-08 & 3 & 13 (6-1-1-5) & AdaEnsemble & 96.94 \% & 94.12 \% & 100 \% & 90.53 \% \\
\hline
2020-12-09 & 1 & 17 (6-1-1-9) & AdaEnsemble & 86.36 \% & 100 \%  & 100 \% & 90.68 \% \\
\hline
2020-12-09 & 2 & 12 (6-1-1-4) & Kalman & 91.26 \% & 94.12 \% & 84.21 \% & 75 \% \\
\hline
2020-12-14 & 1 & 18 (6-1-1-10) & AdaEnsemble & 96 \% & 100 \%  & 94.12 \% & 95.83 \% \\
\hline
2020-12-14 & 2 & 16 (6-1-1-8) & Kalman & 96.97 \% & 88.89 \% & 83.33 \% & 75.15 \% \\
\hline
2020-12-15 & 2 & 17 (6-1-1-9) & AdaEnsemble & 100 \% & 100 \%  & 100 \% & 88.3 \% \\
\hline
2020-12-15 & 3 & 15 (6-1-1-7) & Kalman & 92.23 \% & 100 \% & 100 \% & 80.15 \% \\
\hline
2020-12-16 & 1 & 15 (6-1-1-7) & Kalman & 95.05 \% & 100 \%  & 100 \% & 92.44 \% \\
\hline
2020-12-16 & 2 & 15 (6-1-1-7) & AdaEnsemble & 96.97 \% & 100 \% & 100 \% & 94.02 \% \\
\hline
\end{tabular}
\begin{flushleft}
{\color{red}
* The numbers in the parentheses indicate the number of blocks for each task. The numbers are in the form of (T-B-S-R), where T,B,S, and R are the block numbers for training, test with radial-8-big, test with radial-8-small, test with random target pursuit task, respectively.}
\end{flushleft}
\end{table*}

\begin{table*}[h]
\renewcommand{\arraystretch}{1.3}
\caption{Data details of session in comparison with single models.}
\label{tab:compare_2}
\centering
\begin{tabular}{c|c|c|c|c|c|c|c}
\hline
Date & Session &Block& Decoder & Train Acc & Radial-8 / Big Acc & Radial-8 / Small Acc & \modify{RTP / Small Acc}\\
\hline
2020-12-21 & 1 & 15 (6-1-1-7)* & AdaEnsemble & 93.2 \% & 94.12 \%  & 100 \% & 88.19 \% \\
\hline
2020-12-21 & 2 & 13 (6-1-1-5) & Linear & 77.78 \% & 88.89 \% & 100 \% & 79.8 \% \\
\hline
2020-12-22 & 1 & 12 (6-1-1-4) & AdaEnsemble & 94.12 \% & 83.33 \%  & 100 \% & 84.38 \% \\
\hline
2020-12-22 & 2 & 11 (6-1-1-3) & NN & 78.63 \% & 100 \% & 94.12 \% & 94.12 \% \\
\hline
2020-12-22 & 3 & 12 (6-1-1-4) & Linear & 96 \% & 84.21 \%  & 100 \% & 94.91 \% \\
\hline
2020-12-29 & 1 & 7 (6-1-0-0) & Linear & 69.05 \% & 50 \% & - \% & - \% \\
\hline
2020-12-29 & 3 & 10 (6-1-1-2) & AdaEnsemble & 85.59 \% & 80 \%  & 94.12 \% & 88.33 \% \\
\hline
2021-01-04 & 1 & 13 (6-1-1-4) & AdaEnsemble & 72 \% & 100 \% & 100 \% & 92.19 \% \\
\hline
2021-01-04 & 2 & 7 (6-1-0-0) & NN & 53.47 \% & 26.67 \%  & - \% & - \% \\
\hline
2021-01-06 & 1 & 12 (6-1-1-4) & AdaEnsemble & 91.43 \% & 100 \% & 100 \% & 95.16 \% \\
\hline
2021-01-06 & 2 & 10 (6-1-1-2) & NN & 84.4 \% & 75 \% & 71.43 \% & 65.31 \% \\
\hline
2021-01-06 & 3 & 11 (6-1-1-3) & Linear & 89.42 \% & 100 \% & 88.89 \% & 77.05 \% \\
\hline
2021-01-07 & 1 & 11 (6-1-1-3) & Linear & 78.95 \% & 100 \%  & 100 \% & 88.68 \% \\
\hline
2021-01-07 & 2 & 5 (5-0-0-0) & NN & 64.44 \% & - \% & - \% & - \% \\
\hline
2021-01-07 & 3 & 12 (6-1-1-4) & NN & 95.05 \% & 100 \% & 88.89 \% & 88.68 \% \\
\hline
2021-01-13 & 1 & 11 (6-1-1-3) & NN & 82.61 \% & 75 \% & 76.19 \% & 88.89 \% \\
\hline
2021-01-13 & 2 & 12 (6-1-1-4) & Linear & 94.12 \% & 94.12 \% & 100 \% & 89.23 \% \\
\hline
2021-01-13 & 3 & 12 (6-1-1-4) & AdaEnsemble & 92.31 \% & 100 \% & 94.12 \% & 96.77 \% \\
\hline

\end{tabular}
\begin{flushleft}
{\color{red}
* The numbers in the parentheses indicate the number of blocks for each task. The numbers are in the form of (T-B-S-R), where T,B,S, and R are the block numbers for training, test with radial-8-big, test with radial-8-small, test with random target pursuit task, respectively.}
\end{flushleft}
\end{table*}

\subsection{Offline Decoding Performance}
\label{sec:offline}

In offline trajectory reconstruction, we use monkey's neural data with hand control tasks because it provides ground truth trajectory for evaluation. Two publicly available neural signal datasets are employed for performance evaluation. 

The first one is the ZJUNDD dataset \cite{zhou2014decoding}. The monkey was trained to perform the four-directional center-out task using a joystick. The neural signals were collected by a 96-channel microelectrode array implanted in the contralateral primary motor cortex. \modify{The dataset contains eight subsets and each lasts for 10 minutes, among which the first 5-minute is for training and the second 5-minute is for test. The neural signals were formed in 100 ms bins, with the corresponding kinematic data.} The details of dataset description and signal acquisition can be found in \cite{zhou2014decoding}. The second one is a Zenodo dataset \cite{o2017nonhuman}. The monkey was trained to complete self-paced reaching tasks with a square grid. Neural data were recorded from the primary motor cortex area of a monkey using a 96-channel microelectrode array. Hand velocities were obtained from the position by using a discrete derivative. The recording duration for the data (20170124 01) was about 10 minutes. 

We compare the offline decoding performance of AdaEnsemble with the velocity Kalman filter. For both datasets, the neural signals and kinematic data are normalized by z-score. \modify{For the ZJUNDD dataset, we use the first 2500 samples to train the neural networks and the last 500 samples as a validation set. The smoothing window size for neural signals is 6. Both NN-1 and NN-2 have one hidden layer, and their neuron numbers are set to 16 and 32, respectively. The neuron numbers of the output layer will be adjusted according to the neural dimension of the subset. For the Zenedo dataset, the performance is evaluated with five-fold cross-validation. Each training set has about 3539 samples. The smoothing window size for neural signals is set to 3.  Both NN-1 and NN-2 has two hidden layers, and their neuron numbers are set to (16, 32) and (64, 128), respectively. The neuron number of the output layer is 96, which is consistent with the neuron dimension.} The offline decoding performance is presented in Table \ref{tab:offline}. Overall, AdaEnsemble outperforms the Kalman filter with both datasets. With the ZJUNDD dataset, the decoding CC of AdaEnsemble is 0.859, which is about 11\% higher than the CC of Kalman filter. With the Zenodo dataset, AdaEnsemble achieves CC of 0.858, which is about 15\% higher than the CC of the Kalman.

\begin{figure}[h]
\centering
\includegraphics[scale=0.15]{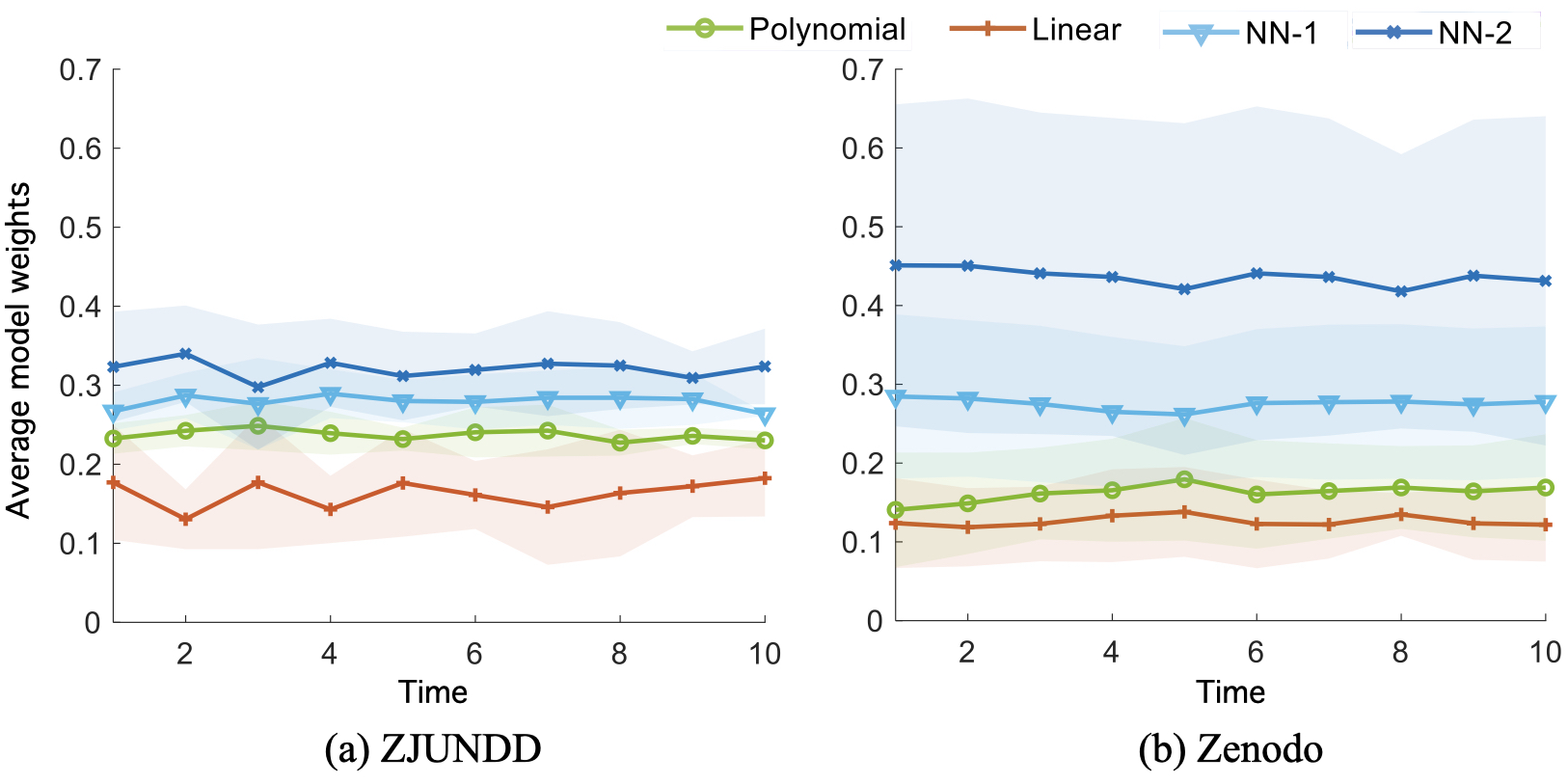}
\caption{{Illustration of model weights in AdaEnsemble with offline decoding. } 
}
\label{fig:offline-weights}
\end{figure}

We further analyze the model weights of AdaEnsemble in offline prediction. As illustrated in Fig. \ref{fig:offline-weights}, for both datasets, the neural network with a larger parameter size obtains the highest weights, while the linear model has the lowest weights. In performance comparison of single models, we find that for both datasets NN-2 achieves the best CC.

\begin{figure}[h]
\centering
\includegraphics[scale=0.15]{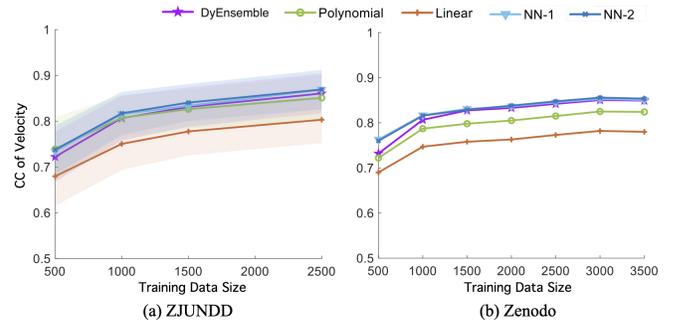}
\caption{{Offline model performance comparison in different training data size. } 
}
\label{fig:data-size-CC}
\end{figure}

\begin{figure}[h]
\centering
\includegraphics[scale=0.15]{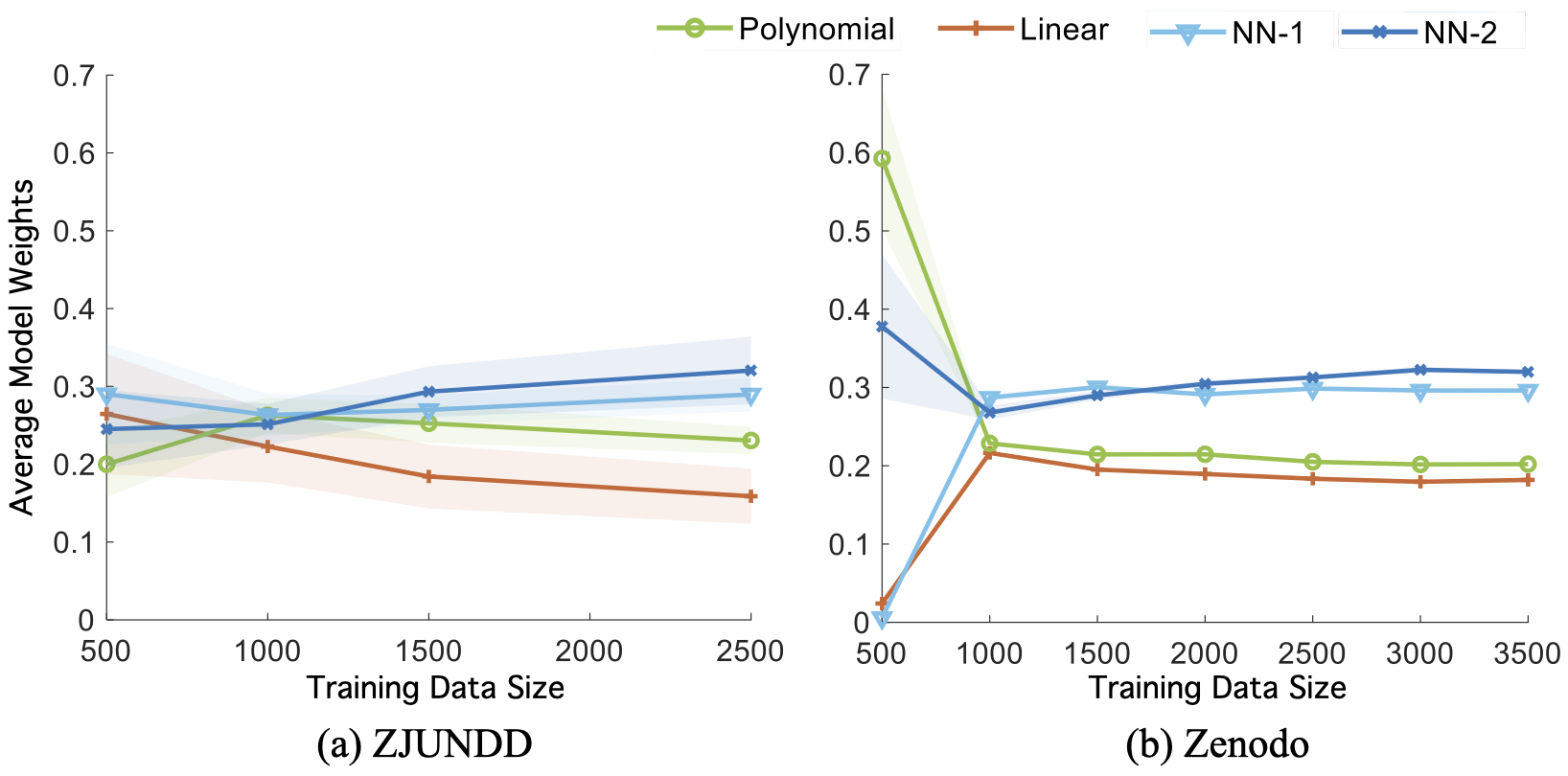}
\caption{{Offline model weights comparison in different training data size. } 
}
\label{fig:data-size-Weights}
\end{figure}

To fairly compare the online and offline situation, we further evaluate the offline performance using similar data sizes with online settings. In the online experiment, there are about 500 points for each block. Therefore, in the offline experiment, we evaluate with sample sizes of 500 to 3500, for comparison with online experiments. In Fig. \ref{fig:data-size-CC}, we find that with the single models, the two neural networks obtain the best performance while the linear model performs the worst. In Fig. \ref{fig:data-size-Weights}, the two neural networks obtained the highest weights, while the linear model obtained the lowest weights. Results demonstrate that nonlinear decoders outperform linear ones with larger datasets, while the performance of AdaEnsemble is among the top decoders.

\begin{table}[tb]
\renewcommand{\arraystretch}{1.3}
\caption{Offline decoding performance with two datasets.}
\label{tab:offline}
\centering
\begin{tabular}{c|cc|ccccc}
\hline
\multirow{2}{*}{Model}  & \multicolumn{2}{c|}{ZJUNDD dataset} & \multicolumn{2}{c}{Zenodo dataset}  \\
 \cline{2-5} & CC & MSE & CC & MSE\\
\hline
Kalman & 0.774$\pm$0.040 & 0.450$\pm$0.003&0.746$\pm$0.011&0.477$\pm$0.001\\
AdaEn & 0.859$\pm$0.044 & 0.367$\pm$0.006&0.858$\pm$0.009&0.269$\pm$0.002\\
\hline 
\end{tabular}
\end{table}

{\color{red}
One interesting observation is that, although nonlinear models demonstrate better performance and obtain higher weights in AdaEnsemble in open-loop decoding (see Supplementary Materials Section II-B), in online BMI control, linear models are with higher weights. It might be due to the feedback-related difference between open- and closed-loop systems \cite{koyama2010comparison, dangi2013design}.
A possible hypothesis is that, since linear models provide more intuitive control, it is easier for the participant to learn from errors of linear models and compensate for them. However, experiments with more subjects should be carried out for further evaluations. In both online and offline situation, AdaEnsemble is able to select more optimal models with higher weights in the assembling, which demonstrate the effectiveness of the adaptive model adjustment. 
}
\subsection{Details of Online Experiment Sessions}

Online experiments are carried out to compare AdaEnsemble with Kalman filter, and single models. The comparison of AdaEnsemble with Kalman filter took 5 experiment days as specified in Table \ref{tab:compare_1}. The comparison with single models took 7 experiment days as specified in Table \ref{tab:compare_2}. On each experiment day, there were several sessions to evaluate different decoders respectively, and the order was randomly assigned. For comparison with single models, an experiment day can contain two or three sessions according to the willingness of the participant, thus there are only two decoders evaluated at a day. While overall, all the decoders are evaluated with six independent sessions. For Linear and NN models, there are three incomplete sessions where we aborted manually due to bad control performance, and these sessions are excluded in performance comparison. It is still worth noting that, the incomplete sessions indicate unstable performance with single models.

\ifCLASSOPTIONcaptionsoff
  \newpage
\fi

\bibliographystyle{IEEEtran}
\bibliography{tbme}
